\documentclass[runningheads]{llncs}
\usepackage[T1]{fontenc}
\usepackage{graphicx}
\usepackage{booktabs}
\usepackage[misc]{ifsym}

\usepackage{mwe}
\usepackage{graphicx}
\usepackage{subcaption}
\usepackage{amsmath}
\usepackage{caption} 
\usepackage{amssymb}
\usepackage{bm}
\usepackage{rotating}

\usepackage{pifont}
\usepackage{siunitx}
\newcommand{\cmark}{\ding{51}}%
\newcommand{\xmark}{\ding{55}}%
\usepackage{adjustbox}
\usepackage{arydshln}

\usepackage{multirow}
\usepackage{xparse}
\usepackage{etoolbox}
\newrobustcmd\B{\DeclareFontSeriesDefault[rm]{bf}{b}\bfseries}  
\usepackage{pdflscape}
\PassOptionsToPackage{hyphens}{url}
\usepackage{hyperref}
\usepackage[perpage]{footmisc}
\usepackage{placeins}
\usepackage{xcolor}
\definecolor{figred}{RGB}{198, 100, 38}
\definecolor{figblue}{RGB}{48, 113, 173}

\begin{document}

\title{From Pixels to Graphs: Deep Graph-Level Anomaly Detection on Dermoscopic Images}

\titlerunning{Deep Graph-Level Anomaly Detection on Dermoscopic Images}

\author{Dehn Xu\inst{1}\orcidID{0009-0005-8457-8153} \and
Tim Katzke\inst{1,2}\orcidID{0009-0000-0154-7735} \and
Emmanuel Müller\inst{1,2}\orcidID{0000-0002-5409-6875}}

\authorrunning{D. Xu et al.}

\institute{Department of Computer Science, TU Dortmund University, Germany \and
Research Center Trustworthy Data Science and Security, University Alliance Ruhr (UA Ruhr), Germany \\
\email{\{dehn.xu,tim.katzke\}@tu-dortmund.de}}

\maketitle              

\begin{abstract}

Graph Neural Networks (GNNs) have emerged as a powerful approach for graph-based machine learning tasks. 
Previous work applied GNNs to image-derived graph representations for various downstream tasks such as classification or anomaly detection. These transformations include segmenting images, extracting features from segments, mapping them to nodes, and connecting them. 
However, to the best of our knowledge, no study has rigorously compared the effectiveness of the numerous potential image-to-graph transformation approaches for GNN-based graph-level anomaly detection (GLAD).
In this study, we systematically evaluate the efficacy of multiple segmentation schemes, edge construction strategies, and node feature sets based on color, texture, and shape descriptors to produce suitable image-derived graph representations to perform graph-level anomaly detection.
We conduct extensive experiments on dermoscopic images using state-of-the-art GLAD models, examining performance and efficiency in purely unsupervised, weakly supervised, and fully supervised regimes.
Our findings reveal, for example, that color descriptors contribute the best standalone performance, while incorporating shape and texture features consistently enhances detection efficacy. In particular, our best unsupervised configuration using OCGTL achieves a competitive AUC-ROC score of up to $0.805$ without relying on pretrained backbones like comparable image-based approaches. With the inclusion of sparse labels, the performance increases substantially to $0.872$ and with full supervision to $0.914$ AUC-ROC.
\end{abstract}

\section{Introduction}

Anomaly detection in images plays a pivotal role across a wide range of applications, from identifying cracks or surface defects in industrial inspection~\cite{bergmannMVTecADComprehensive2019} to spotting unusual activities in security footage~\cite{duongDeepLearningBasedAnomalyDetection2023}, and detecting tumors or lesions in medical imaging~\cite{caiComparativeStudyOfAnomalyDetectionInMedicalImages2025}. State-of-the-art convolutional neural networks and vision transformers excel at these tasks when large, labeled datasets are available. However, they treat every pixel uniformly and often require extensive pretraining. In anomaly detection, where normal examples vastly outnumber the rare, unpredictable anomalies, this pixel-level redundancy can obscure subtle deviations, inflate computational and data availability costs, and demand powerful hardware that may be impractical in resource-limited clinical or edge settings.

Graph‐structured representations offer a compelling alternative by abstracting an image into a compact set of nodes and edges~\cite{cosmaGeometricSuperpixelRepresentations2023,feyConvolutionalNeuralNetworks2017,rodriguesGraphConvolutionalNetworks2024}. Nodes correspond to locally homogeneous regions, defined by superpixels, learned patches, or provided segmentation masks, and can be associated with higher-level features, abstracted from these regions' low-level pixel content. Edges capture spatial adjacency, feature similarity, or both. For anomaly detection, this compressed representation sharply reduces input dimensionality and noise, enabling models to focus on semantically meaningful units rather than millions of individual pixels. The relational inductive bias of graphs likewise confers robustness to small rotations or translations, ensuring that a tiny shift in position does not mask critical deviations without needing data augmentation. Moreover, their compactness allows training lightweight graph-anomaly detectors from scratch, eschewing massive pretrained backbones, thereby reducing both runtime and energy consumption. 

\begin{figure}[tbp!]
    \centering
    \includegraphics[width=\textwidth]{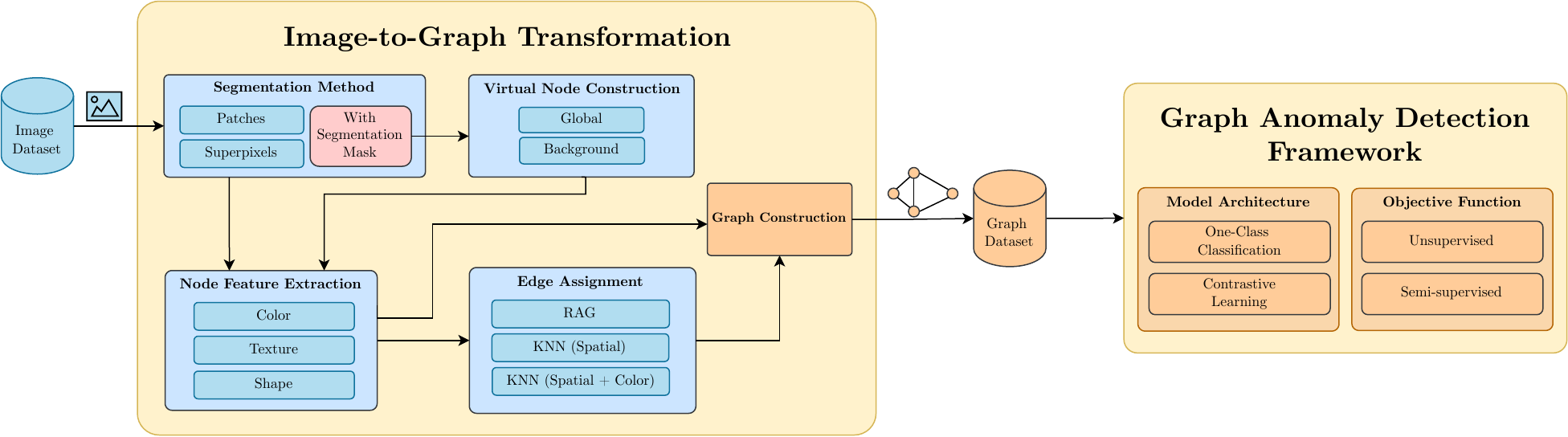}
    \caption[Methodology of graph transformation and GAD]{Overall process of our two-step evaluation study. Images are transformed into graph representations, on which we conduct graph anomaly detection.}
    \label{fig:approach}
\end{figure}

In this work, we explore these potential advantages in conjunction with raw anomaly detection performance systematically and build a bridge between image-based and graph-based anomaly detection. To ground this in a high-stakes medical scenario, we focus on the HAM10000 image dataset~\cite{tschandlHAM10000DatasetLarge2018}. This corpus of high-resolution skin-lesion images contains a clear majority of benign nevi as natural ``normal'' samples alongside multiple smaller, clinically significant classes as diverse anomalies (melanoma, basal cell carcinoma, vascular lesions, and others) and provides pixel-level segmentation masks. Crucially, node features, like shape descriptors, texture histograms, size measures, and color moments, directly align with the established ABCDE criteria~\cite{friedman1985early} (Asymmetry, Border irregularity, Color variation, Diameter, Evolving) used by dermatologists. As is most common in anomaly detection, we evaluate our pipelines primarily in the unsupervised setting (training exclusively on nevi examples) while also investigating supervised variants that leverage varying degrees of labeled anomalies.

As our main contribution, we systematically analyze the effectiveness of various combinations of segmentation strategies, edge-construction methods, node-feature sets, and state-of-the-art graph anomaly detection algorithms in the aforementioned context. By isolating each component, we quantify how region compression, relational robustness, and model compactness translate into detection performance, data reduction, as well as training and inference speed. We focus on identifying bottlenecks in each step of the process, highlighting areas for improvement and future research. Our findings reveal that comparable performance to the results of other image-based studies can still be achieved, even with the reduction of the available data features via segmentation. Furthermore, this data reduction has distinct advantages in both runtime and data efficiency. To ensure reproducibility, our implementation is publicly available at \url{https://github.com/deX-de/Deep-GLAD-on-Dermoscopic-Images}.

\section{Related Work} 
Researchers have applied the concept of converting images into graphs in various machine learning studies.
Han et al.~\cite{hanVisionGNNImage2022} proposed an end-to-end approach that transforms images into non-overlapping patches, extracts features using a CNN-stem, and trains on a GNN.
Most other works employed a two-step approach with a separate image-to-graph transformation and downstream task. Specifically, they focused on superpixel algorithms such as SLIC~\cite{achantaSLICSuperpixelsCompared2012}, Quickshift~\cite{vedaldiQuickShiftKernel2008} or Felzenszwalb~\cite{felzenszwalbEfficientGraphBasedImage2004} to first segment the image into meaningful regions, extract relevant features from these regions and assign edges between the nodes~\cite{cosmaGeometricSuperpixelRepresentations2023,feyConvolutionalNeuralNetworks2017,rodriguesGraphConvolutionalNetworks2024}. Subsequently, they utilized GNNs to learn on extracted image-derived graph representations.
The study conducted by Annaby et al.~\cite{annabyMelanomaDetectionUsing2021a} converted images into graphs in the context of melanoma classification.
Similarly, they transformed the image into a region adjacency graph using the SLIC algorithm.
However, in their downstream task, they handcrafted graph- and node-level features in the spatial and spectral domain. They then used these features in shallow machine learning models in combination with conventional image-based features.

Graph anomaly detection (GAD) seeks to uncover irregularities or non-conformity in graph-structured data at varying granularities~\cite{qiao2025deep}. Usually, this is done in an unsupervised or weakly supervised setting. Principal methodological paradigms encompass reconstruction-based models that learn to regenerate normal graph elements and flag high reconstruction error as anomalous~\cite{luo2022deep,niu2023graph}; contrastive techniques that derive normality by discriminating between augmented or heterogeneous views~\cite{liSimplifiedTransformer2023,liuSelfInterpretableGraphlevelAnomalyDetection2023,ma2022deep}; and one-class classification approaches that enclose normal instances within a compact decision region, treating outliers beyond its boundary as anomalies~\cite{qiuRaisingBarGraphlevel2022,zhaoUsingClassificationDatasets2021}.
For GAD on image-derived graph representation, the aim is either (1) detecting unusual objects or patterns in an image (essentially node-level anomaly detection)~\cite{acharya2022detecting,tu2022hyperspectral,tu2024anomaly,wang2023multi}, (2) detecting if a whole image is anomalous (graph-level anomaly detection)~\cite{zoghlami2024viglad}, or both~\cite{gu2025univad,peng2025sam,xie2023pushing}. 
While GAD on image-derived graphs is not a new phenomenon, a thorough analysis of the specific performances of diverse candidate combinations of image-to-graph transformations with current state-of-the-art GLAD methods is missing.

\section{Preliminaries and Problem Definition}\label{problem_definition}

A graph $\mathcal{G}$ is an ordered pair $\mathcal{(V, E)}$ of $n=|\mathcal{V}|$ nodes forming the \emph{node set} $\mathcal{V}=\{v_1,\ldots,v_n\}$ 
that can be considered as abstract representations of entities together with the \emph{edge set}
$\mathcal{E}\subseteq \mathcal{V} \times \mathcal{V}$ that show the relationships between aforementioned entities. 
Two nodes $u, v \in \mathcal{V}$ are \emph{adjacent} if $(u, v) \in \mathcal{E}$ or $(v, u) \in \mathcal{E}$. 
An \emph{attributed graph}, represented as $\mathcal{G} = (\mathcal{V, E}, X)$, includes an attribute set $X =\{\mathbf{x}_v \in \mathbb{R}^d \, | \, v\in \mathcal{V}\}$, where $d$ is the node feature dimension.
In principle, an image $\mathbf{I}\in \mathbb{R}^{W\times H\times C}$ of width $W$, height $H$, and color channels $C$, can be thought of as a $4 \cdot D$-connected graph with $\mathcal{V}=\{v_i \,|\, i \in [W\cdot H]\}$, where $\phi:\mathcal{V}\to [W]\times [H]$ is a bijective mapping from nodes to pixel coordinates, $\mathcal{E}=\{(v_i, v_j\} \,|\, v_i\neq v_j, \| \phi(v_i) - \phi(v_j)\| \le \sqrt{D}\}$, and $X=\{\mathbf{I}_{x,y,\cdot} \,|\, (x,y) \in [W]\times [H]\}$.

The task of graph-level anomaly detection (GLAD) is to distinguish anomalous graphs from normal ones within a given graph dataset~\cite{maComprehensiveSurveyGraph2023}. Traditionally, researchers used graph kernels to extract graph-level features. They subsequently applied general shallow anomaly detection methods (e.g., Local Outlier Factor (LOF)~\cite{breunigLOFIdentifyingDensitybased2000} or One-Class Support Vector Machines (OC-SVM)~\cite{scholkopfSupportVectorMethod1999}) to detect anomalous graphs. With advances in deep learning and graph representation learning, more sophisticated GNN-based methods automatically extract relevant graph features end-to-end. These models primarily operate unsupervised, with most training data containing only the normal class. However, specific GLAD model architectures can incorporate semi-supervision through minor adjustments to the objective function.

The problem statement of our work can be formalized as follows: Given the image set $\mathcal{D}_{img}=\{(\mathbf{I}_1,y_1),\ldots,(\mathbf{I}_n,y_n)\}$, 
we evaluate different graph construction configurations composed of various segmentation methods, node features, and edge assignments to construct graphs $\mathcal{D}_{graph}=\{(\mathcal{G}_1,y_1), \ldots, (\mathcal{G}_n, y_n)\}$ in the context of anomaly detection. 
For anomaly detection, the labels $y_1,\ldots,y_n$ denote whether a sample is either normal ($y=0$) or anomalous ($y=1$).
The evaluation of these configurations considers various state-of-the-art GLAD methods trained on the training split $\mathcal{D}_{graph}^{train}\subset \mathcal{D}_{graph}$ to predict whether unseen graphs are normal or anomalous.
In an unsupervised setting, we only consider normal data during training. For (semi-)supervised learning, additional labeled anomalies are incorporated into $\mathcal{D}_{graph}^{train}$.

\section{Benchmark Design}\label{sec:benchmark_design}
In this section, we design a two-step graph-based benchmark pipeline tailored to skin lesion analysis on the HAM10000 dataset~\cite{tschandlHAM10000DatasetLarge2018}. Beginning with a clinically relevant, high-resolution dataset, we convert dermoscopic images into graphs through segmentation-based node construction, rich visual feature extraction, edge assignments, and optionally virtual nodes. This representation allows for effective GLAD, which we evaluate using recent state-of-the-art models. Our benchmark is designed to grasp irregular, heterogeneous patterns of dermatological anomalies and allows both unsupervised and semi-supervised detection settings.
\subsection{Dataset Description}
\begin{table}[tbp!]
    \centering
    {%
      \setlength{\tabcolsep}{6pt}%
      \begin{tabular}{cccccccc}
        \toprule
         akiec & bcc & bkl & df & mel & nv & vasc & Total \\
        \midrule 
        327 & 514 & 1099 & 115 & 1113 & 6705 & 142 & 10015 \\
        \bottomrule
      \end{tabular}
    }
    \caption[HAM10000 class distribution \cite{tschandlHAM10000DatasetLarge2018}]{Image distribution of HAM10000.}
    \label{tab:ham10000_distribution}
\end{table}

The Human Against Machine dataset (HAM10000) consists of $\num{10015}$ dermoscopic high-resolution skin lesion images, each with respective diagnoses that are benign or malignant. Correctly diagnosing skin lesions through machine learning is both clinically and economically meaningful, as it makes saving lives possible with fewer human resources. Moreover, the HAM10000 dataset includes segmentation masks, hand-drawn by a professional dermatologist, that match each skin lesion.

Though the availability of specific diagnosis labels facilitates supervised classification, the imbalanced nature of this dataset, as shown in Table~\ref{tab:ham10000_distribution}, with the majority of images belonging to the nevus (nv) class, makes it well-suited for anomaly detection~\cite{caiComparativeStudyOfAnomalyDetectionInMedicalImages2025}. 
Additionally, labeling skin lesions demands the expertise of dermatologists, which involves considerable time and capital. Therefore, HAM10000 is particularly relevant in an unsupervised and semi-supervised context.
Furthermore, the ABCDE schema~\cite{friedman1985early}, used to differentiate between benign and malignant skin lesions, motivates the application of meaningful node features given by their color, shape, and texture.

HAM10000 shares structural homogeneity across all classes. Unlike natural image datasets, where objects possess distinct geometric features, dermatological lesions appear as blob-like regions without a consistent structure. The irregular structure in skin lesions reduces the importance of position encoding, which is often critical in image-based classification. 
This characteristic aligns well with GNNs on graph-level tasks, which naturally handle nodes permutation-invariantly.

\subsection{Image-to-Graph Transformation}\label{sec:image-to-graph}

\begin{figure}[tbp!]
    \centering
    \subfloat[]{
        \centering
        \includegraphics[width=.9\textwidth]{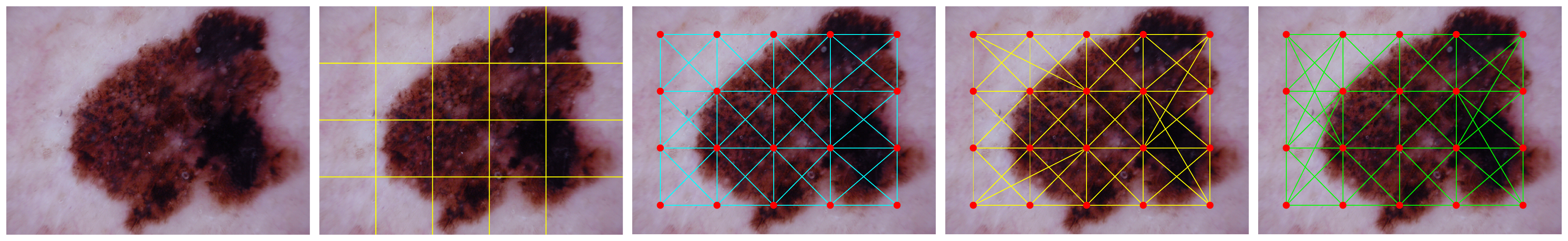}
        \label{fig:ham10000_patch}
    }

    \subfloat[]{
        \centering
        \includegraphics[width=.9\textwidth]{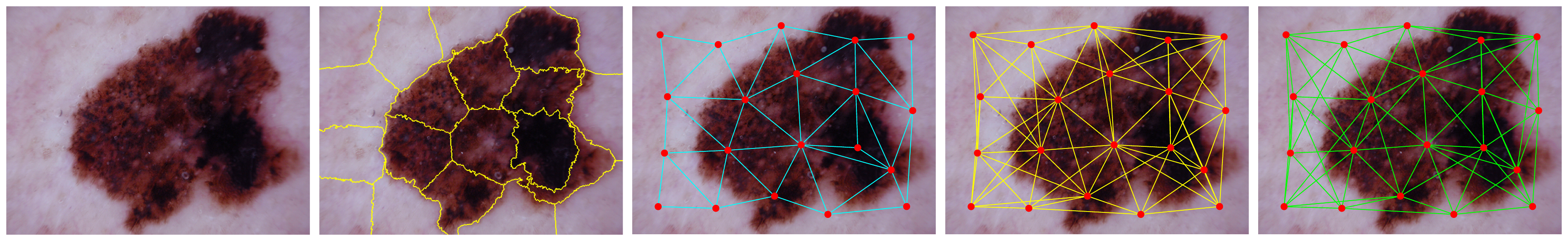}
        \label{fig:ham10000_slico}
    }
    \caption[Visualization of HAM10000 \cite{tschandlHAM10000DatasetLarge2018}]{HAM10000 melanoma image, segmented (\subref{fig:ham10000_patch}) as patches and (\subref{fig:ham10000_slico}) with SLICO, followed by edge construction via RAG, $\text{KNN}_{\text{s}}$ and $\text{KNN}_{\text{sc}}$.
    }
    \label{fig:ham10000_visualization}
\end{figure}

The intuitive image-to-graph transformation, as detailed in Section~\ref{problem_definition}, is generally impractical due to the sheer amount of pixels in high-resolution images.

Hence, the solution explored in our approach is to partition the image into segments of pixels with a segmentation algorithm such as Patch-based decomposition, as visualized in Figure~\ref{fig:ham10000_patch}. With this method, the image is divided into a grid of non-overlapping patches $\mathcal{S} = \{ S_i \mid i = 1, 2, \ldots, n \}$, where each patch $S_i$ is a rectangle of size $P \times P$ pixels with $\lfloor H /P \rfloor\, \cdot \, \lfloor W / P\rfloor = n$. 
Another technique is using a superpixel algorithm, specifically Simple Linear Iterative Clustering (SLIC)~\cite{achantaSLICSuperpixelsCompared2012}, to segment images, commonly in the context of graph classification on images~\cite{rodriguesGraphConvolutionalNetworks2024}. This algorithm assigns pixel coordinates $(x,y)\in [W]\times [H]$ to segments $S_i$ through local $k$-means over a five-dimensional vector space of the pixels' color and spatial features.
We utilize a variant of SLIC, SLIC-zero (SLICO, see Figure~\ref{fig:ham10000_slico}), which dynamically chooses the compactness parameter that controls the weight between the color and spatial distance for each superpixel.

The resulting segments $\mathcal{S}=\{S_1,\ldots,S_n\}$ can then be used to construct the nodes of the graph $\mathcal{V}=\{v_1, \ldots, v_n\}$. 
Subsequently, different edge construction techniques are used to connect these nodes.
The Region Adjacency Graph (RAG) is built analogously to the grid graph. For any two pixel coordinates $(x,y), (x',y')$ of differing segments $S_i, S_j$, and connectivity $D$, if $\|(x,y)-(x',y')\|\le \sqrt{D}$ we connect the nodes $v_i$ and $v_j$.
A connectivity of $1$ and $2$ leads to 4-connected and 8-connected neighborhoods, respectively. Assuming pre-computed node features from the feature extraction process, a more efficient approach is using $k$-nearest neighbors (KNN) on features such as spatial centroid coordinates ($\text{KNN}_{\text{s}}$) or a combination of these coordinates with mean color values ($\text{KNN}_{\text{sc}}$). For our work, all edges are set to be undirected.

Descriptive features are extracted from the image segments to enable comprehensive learning of visual patterns. 
These features transform the raw pixel values of segmented regions into a concise set of numerical attributes. 
We categorize relevant
visual information into three types, which are evaluated both independently and in combination: color, texture, and shape.
For color-based node features, we assign each node the mean color, standard deviation, and skewness of the pixel intensities within its segments in the RGB, HSV, and CIELAB color spaces. 
Texture features are derived from the Local Binary Pattern (LBP)~\cite{ojalaMultiresolutionGrayscaleRotation2002} with $P=8, R=1$ and the Gray-level Co-occurrence Matrix (GLCM)~\cite{haralickTexturalFeaturesImage1973} on contrast, dissimilarity, energy, correlation, and homogeneity at angles $\{\ang{0}, \ang{45}, \ang{90}, \ang{135}\}$. 
Moreover, we extracted a total of $38$ translation-, scaling-, and rotation-invariant moments proposed by~\cite{feyConvolutionalNeuralNetworks2017} as shape features. 

\begin{figure}[tbp!]
    \centering 
    \includegraphics[width=.45\textwidth]{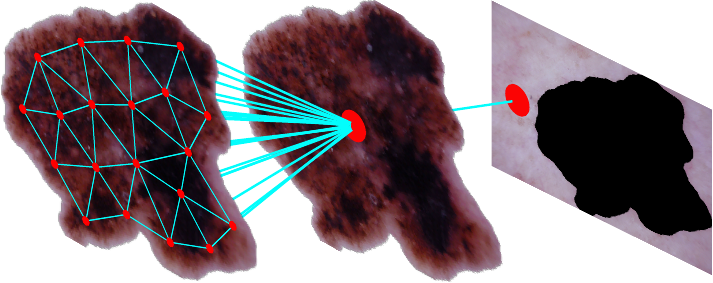}
    \caption[Visualization of virtual nodes used on HAM10000 \cite{tschandlHAM10000DatasetLarge2018}]{Virtual nodes used on HAM10000. 
    Segmented nodes are each connected to the lesion node, which in turn is connected to the skin node.}
    \label{fig:virtual_nodes}
\end{figure}
Since the HAM10000 dataset includes bitmasks, isolation of skin lesions is feasible. This process is clinically relevant due to its simplified ABCDE schema application. However, it also results in the loss of information from the surrounding skin tissue. To address this limitation, we introduce two virtual nodes: $v_g$ and $v_b$. Node $v_g$ represents the isolated skin lesion, while node $v_b$ represents the surrounding skin tissue. Both virtual nodes possess an identical number of node features as those derived from the image segmentation method. As illustrated in Figure~\ref{fig:virtual_nodes}, node $v_g$ is connected to all other nodes, serving as a global communication hub. In contrast, node $v_b$ is linked only to $v_g$.

\subsection{Graph Anomaly Detection Methods}\label{sec:gad_methods}
Next, we evaluated the three best-performing state-of-the-art GLAD models from the most recent graph-level anomaly detection benchmark \cite{wangUnifyingUnsupervisedGraphlevelAnomaly2025}, namely SIGNET \cite{liuSelfInterpretableGraphlevelAnomalyDetection2023}, CVTGAD \cite{liSimplifiedTransformer2023}, and OCGTL \cite{qiuRaisingBarGraphlevel2022} on the obtained graph representations.
SIGNET is a contrastive learning-based approach that optimizes bottleneck subgraphs by enhancing shared structural information across graph views while suppressing irrelevant details. Similarly, CVTGAD extends contrastive learning to both node and graph levels using a simplified transformer with cross-view attention. Finally, OCGTL is an ensemble of $K+1$ feature extractors with a two-part objective function. Each normal graph embedding is mapped to a minimal hypersphere through the one-class classification objective $\mathcal{L}_{\text{OCC}}(\mathcal{G})=\sum_{k=1}^K\|\text{GIN}_k(\mathcal{G})-c\|$, while the contrastive loss $\mathcal{L}_{\text{GTL}}(\mathcal{G})$ between embeddings of $\text{GIN}_{\text{0}}$ and $\{\text{GIN}_k\}_{k=1}^K$ ensures relevance with diversity.

Motivated by~\cite{ruffDeepSemiSupervisedAnomaly2020}, we similarly extend the one-class classification objective of OCGTL to enable a semi-supervised approach to subsequently evaluate the impact of (weak) supervision.
Given the original loss function in \cite{qiuRaisingBarGraphlevel2022}:
\[
\mathcal{L}_{\text{OCGTL}} = \mathbb{E}_{\mathcal{G}}[\mathcal{L}_{\text{OCC}}(\mathcal{G})+\mathcal{L}_{\text{GTL}}(\mathcal{G})],
\]
where $\mathbb{E}_{\mathcal{G}}[\cdot]$ is the expectation over the distribution of $\mathcal{G}$, we modify it such that training maximizes the distance between labeled anomalies and the center:
\begin{align*}
\mathcal{L}_{\text{Semi-OCGTL}}&=\begin{cases}
    \mathbb{E}_{\mathcal{G}}[\sum_{k=1}^K \|\text{GIN}_k(\mathcal{G})-c\|^{-1}+\mathcal{L}_{\text{GTL}}(\mathcal{G})], &y=-1 \\
    \mathbb{E}_{\mathcal{G}}[\mathcal{L}_{\text{OCC}}(\mathcal{G})+\mathcal{L}_{\text{GTL}}(\mathcal{G})], &\text{otherwise}
\end{cases}.
\end{align*}
Here, $y\in\{-1,0,1\}$ corresponds to labeled anomalies, unlabeled samples, and labeled normal samples, respectively. 
We apply semi-supervision exclusively to OCGTL, as extending the other models requires modifications to their architecture, e.g., adding a separate classification head, implementing one-class classification, or introducing a reconstruction-based loss term.
\section{Experiments}

To rigorously evaluate the various combinations of image-to-graph transformations and graph-based anomaly detection methods introduced in Section~\ref{sec:benchmark_design}, we first construct two types of graphs on every lesion image, namely patch-based graphs, where each image is divided into a non-overlapping $4\times5$ grid of $20$ patches, as well as superpixel graphs, with SLICO segmentation using $n=20$. Each strategy is evaluated with and without virtual nodes and the provided ground truth segmentation masks.

For both segmentation strategies, we evaluate edge construction via Region Adjacency Graphs (connectivity $=2$) and $k$-nearest-neighbor graphs based on spatial adjacency with and without color similarity ($k=6$). Node features span the categories from Section~\ref{sec:image-to-graph}, from basic color statistics to advanced texture and shape descriptors (see Appendix \ref{app:image_to_graph} for more details on the specifics of the image-to-graph transformations).
All anomaly-detection models introduced in Section~\ref{sec:gad_methods} are evaluated, each with a lightweight backbone of two GIN layers and hidden dimension = $16$, to mitigate the well-known issue of over-smoothing~\cite{ruschSurveyOversmoothingGraphNeural2023} on these small graphs. Details regarding specific hyperparameters are provided in Appendix \ref{app:gad}.

\subsection{Experimental Setup}\label{sec:exp_setup}

We employ five-fold class-stratified cross-validation on the official HAM10000 training set with a fixed random seed. In each split, four folds ($\approx 80 \%$) form the training set, and the remaining fold ($ \approx20\%$) is held out for testing. We frame anomaly detection as one-vs-rest: nevus is ``normal'', with other lesion types (melanoma, basal cell carcinoma, etc.) being ``anomalous''.

For the unsupervised experiments, we trained each model exclusively on nevus samples from the training folds, ignoring the anomalous samples. Evaluation then proceeded on the entire test fold. Motivated by literature indicating that even a small fraction of labeled anomalies can substantially improve performance~\cite{hanADBenchAnomalyDetection2022}, we also explored two supervised regimes applying the semi-supervised adaptation of OCGTL under identical conditions. For weak supervision, we retained random samples of anomalies (in addition to all nevus examples) from the training folds, s.t. the respective training folds comprised of $5\%$ labeled anomalies. For full supervision, we included every labeled anomaly from the training folds alongside the nevus samples. Contrary to classification, we still do not differentiate between anomalous classes.

All models are trained for $20$ epochs with Adam (learning rate = $0.001$) and a batch size of $128$. To counter the curse of dimensionality, we apply PCA to each feature set (except for $\text{RGB}_{\text{avg}}$) prior to training, retaining $95 \%$ of the variance; this reduces computational cost while preserving the informative signal.

\subsection{Performance Comparison}\label{sec:performance}

\begin{table}[tbp!]
    \centering
    \begin{adjustbox}{width=\textwidth}
    \begin{tabular}{llcccccc}
        \toprule
        \multirow{2}{*}{} 
        & \multirow{2}{*}{Features} 
        & \multicolumn{3}{c}{\textsc{Patch}} 
        & \multicolumn{3}{c}{\textsc{SLICO}}  \\
        \cmidrule(lr){3-5}\cmidrule(lr){6-8}
        & 
        &  RAG     & $\text{KNN}_{\text{s}}$       & $\text{KNN}_{\text{sc}}$  
         &  RAG     & $\text{KNN}_{\text{s}}$     & $\text{KNN}_{\text{sc}}$            \\
        \midrule
        \multirow{5}{*}{\shortstack[l]{Mask \xmark \\ VN \xmark}} 
        & $\text{RGB}_{\text{avg}}$ 
            & 70.9$\pm$2.4 
            & \underline{71.8$\pm$2.2} 
            & 66.5$\pm$2.3 
            & 66.8$\pm$3.4 
            & 64.4$\pm$2.0 
            & 64.9$\pm$3.7       \\
        & Color 
            & \underline{\B 74.2$\pm$2.0} 
            & 72.5$\pm$1.8 
            & 70.9$\pm$2.0       
            & 73.0$\pm$1.8 
            & 71.8$\pm$2.0      
            & 72.3$\pm$1.4       \\
        & Texture 
            & 68.3$\pm$2.4      
            & \underline{69.2$\pm$1.6} 
            & 66.9$\pm$1.9     
            & 68.2$\pm$1.8 
            & 67.6$\pm$2.2      
            & 67.5$\pm$1.6       \\
        & Shape 
            & ---               
            & ---               
            & ---               
            & \underline{59.6$\pm$1.2} 
            & 55.1$\pm$1.4      
            & 55.9$\pm$1.7       \\
        & All 
            & 73.8$\pm$3.0 
            & \underline{\B 75.0$\pm$2.3} 
            & 70.9$\pm$1.5 
            & 72.4$\pm$1.6 
            & 71.4$\pm$1.4      
            & 71.2$\pm$2.1       \\
        \midrule 
        \multirow{5}{*}{\shortstack[l]{Mask \cmark \\ VN \cmark}} 
        & $\text{RGB}_{\text{avg}}$ 
            & 68.7$\pm$2.4 
            & 67.7$\pm$3.0 
            & 67.7$\pm$2.0   
            & \underline{72.8$\pm$2.2} 
            & 71.7$\pm$2.5      
            & 70.2$\pm$2.9       \\
        & Color 
            & 69.5$\pm$1.3 
            & 69.3$\pm$1.6 
            & 70.2$\pm$1.5       
            & \underline{\B 80.6$\pm$1.6} 
            & \B 77.5$\pm$1.9      
            & \B 78.6$\pm$1.7     \\
        & Texture 
            & 59.4$\pm$1.6      
            & 59.7$\pm$3.0 
            & 61.3$\pm$2.0     
            & \underline{66.5$\pm$2.0} 
            & 63.6$\pm$2.1      
            & 64.6$\pm$1.6       \\
        & Shape 
            & ---               
            & ---               
            & ---               
            & 59.2$\pm$1.9 
            & 59.7$\pm$1.4      
            & \underline{60.4$\pm$1.5}       \\
        & All 
            & 69.5$\pm$1.8 
            & 71.3$\pm$2.3 
            & \B 71.3$\pm$1.7 
            & \underline{76.9$\pm$1.7} 
            & 75.4$\pm$1.8      
            & 76.6$\pm$1.3       \\
        \bottomrule
    \end{tabular}
    \end{adjustbox}
    \caption[Unsupervised HAM10000 ROC-AUC results]{Mean AUC-ROC and SD values in \% per image-to-graph transformation pipeline across all GLAD models, averaged over all splits. Best performance per column in bold, best performance per row underlined. For the features, 'All' corresponds to the complete feature set.}
    \label{tab:ham10000_roc_auc_2}
\end{table}

In the following, we analyze how image-to-graph transformation pipelines, GLAD models, and supervision levels influence the image-level AUC-ROC performance (reported in \%). Analogous plots and tables for AUC-PR, along with the exact per-configuration results, are provided in Appendix~\ref{app:performance}.  For patch-based segmentation, shape features are excluded, as they are not informative. 

Table~\ref{tab:ham10000_roc_auc_2} summarizes performance by graph-construction pipeline, averaged over all GLAD models and data splits. In this table, combinations with Mask but without VN are omitted, as their performance was generally similar to or worse than the corresponding VN variant; detailed plots including these variants are provided in Appendix~\ref{app:performance}. For SLICO, using both VN and Mask yields better performance, whereas for Patch, the configuration without either performs best. RAG is (bar one exception) the strongest edge-construction strategy with SLICO, with no clear winner for Patch-based segmentation. Node feature choice has a pronounced effect: Color alone is the strongest single feature set (appearing in all three top average configurations) and performs comparably to variants that additionally include Texture and Shape. However, the benefit of adding Texture and Shape depends strongly on supervision and the GLAD model. The complete feature set substantially improves the best unsupervised method (SIGNET) and all supervised OCGTL variants in the absence of virtual nodes and segmentation masks.

\begin{figure}[tp!]
  \centering
  \includegraphics[width=\textwidth]{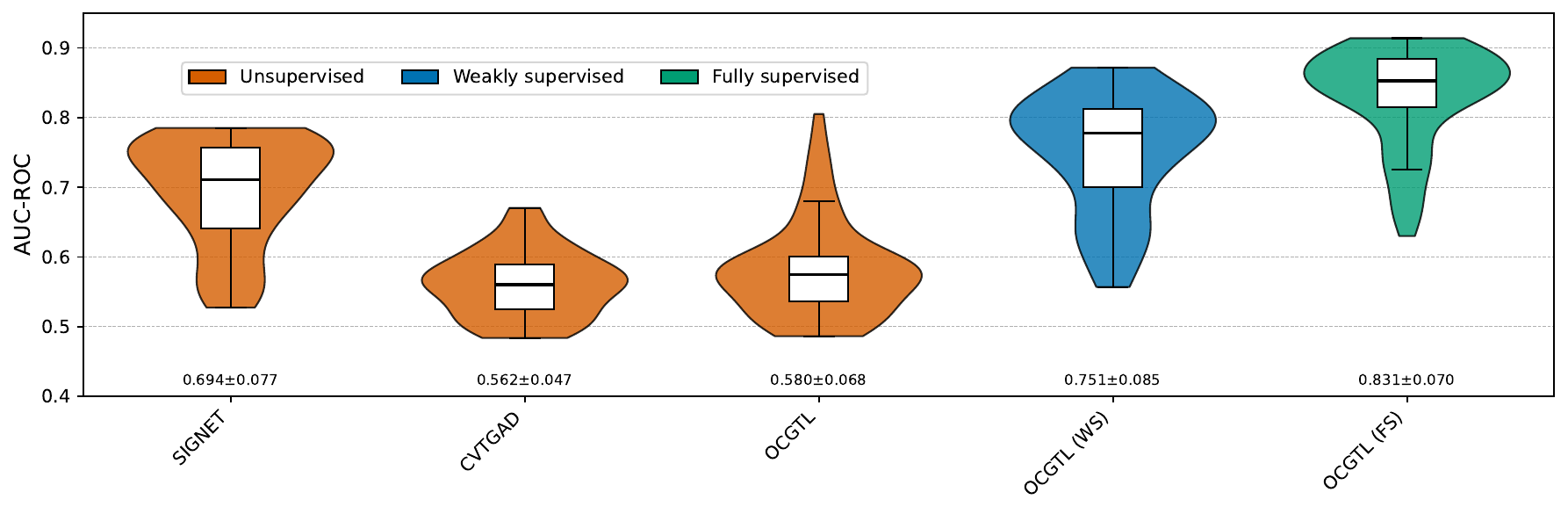}
  \caption{AUC-ROC performance distribution including median, lower quartile, and upper quartile per GLAD model, as well as mean and SD, across all image-to-graph transformations.}
  \label{fig:violin_aucroc}
\end{figure}

Performance differences across GLAD models and supervision regimes are substantial. Figure~\ref{fig:violin_aucroc} shows the distribution of AUC-ROC values across all image-to-graph pipelines and data splits. In the unsupervised setting, SIGNET achieves the strongest average performance, while the overall best result is obtained with OCGTL. With Mask and VN, SIGNET reaches up to $77.4 \pm 1.4$ and OCGTL $80.5 \pm 1.9$; without Mask and VN, the respective values are $72.0 \pm 1.2$ and $68.1 \pm 2.2$. In comparison, CVTGAD attains up to $66.8 \pm 5.1$ with Mask and VN, and $66.4 \pm 3.8$ without either. Notably, with the complete node feature set, SIGNET is relatively insensitive to the segmentation and edge-construction choices. Under weak supervision with only $5\%$ anomalies in the training set, OCGTL improves to $87.2 \pm 0.6$, and in the fully supervised setting it reaches $91.4 \pm 0.2$. In general, the supervised variants benefit less from VN and Mask but more from the complete feature set.

To contextualize these findings, a recent study~\cite{caiComparativeStudyOfAnomalyDetectionInMedicalImages2025} reports image-level AUC-ROC scores for a variety of image-based anomaly-detection methods employing ImageNet-pretrained feature extractor backbones on the ISIC2018 dataset~\cite{isic2018_challenge} (with HAM10000 constituting its training set, and all images originating from the same source). The results are obtained under unified evaluation protocols and averaged over three independent runs per method. The authors adopt the same one-vs-all setup, using nevus as the normal class and the other six classes as anomalies, with training conducted on $6,705$ normal images and testing on $1,512$ images ($909$ normal, $603$ anomalous). Here, the best reconstruction-based method configuration achieves an AUC-ROC of $79.2 \pm 0.6$, while the best self-supervised configuration reaches $80.7 \pm 0.5$. This places our best unsupervised GLAD configuration with Mask and VN (OCGTL with RAG+SLICO+Color, $80.5 \pm 1.9$) within the range of the strongest image-based baselines on a closely comparable task definition.

\subsection{Efficiency Analysis}

\begin{figure}[tbp]
  \centering
  \begin{subfigure}[b]{0.48\linewidth}
    \centering
    \includegraphics[width=\linewidth]{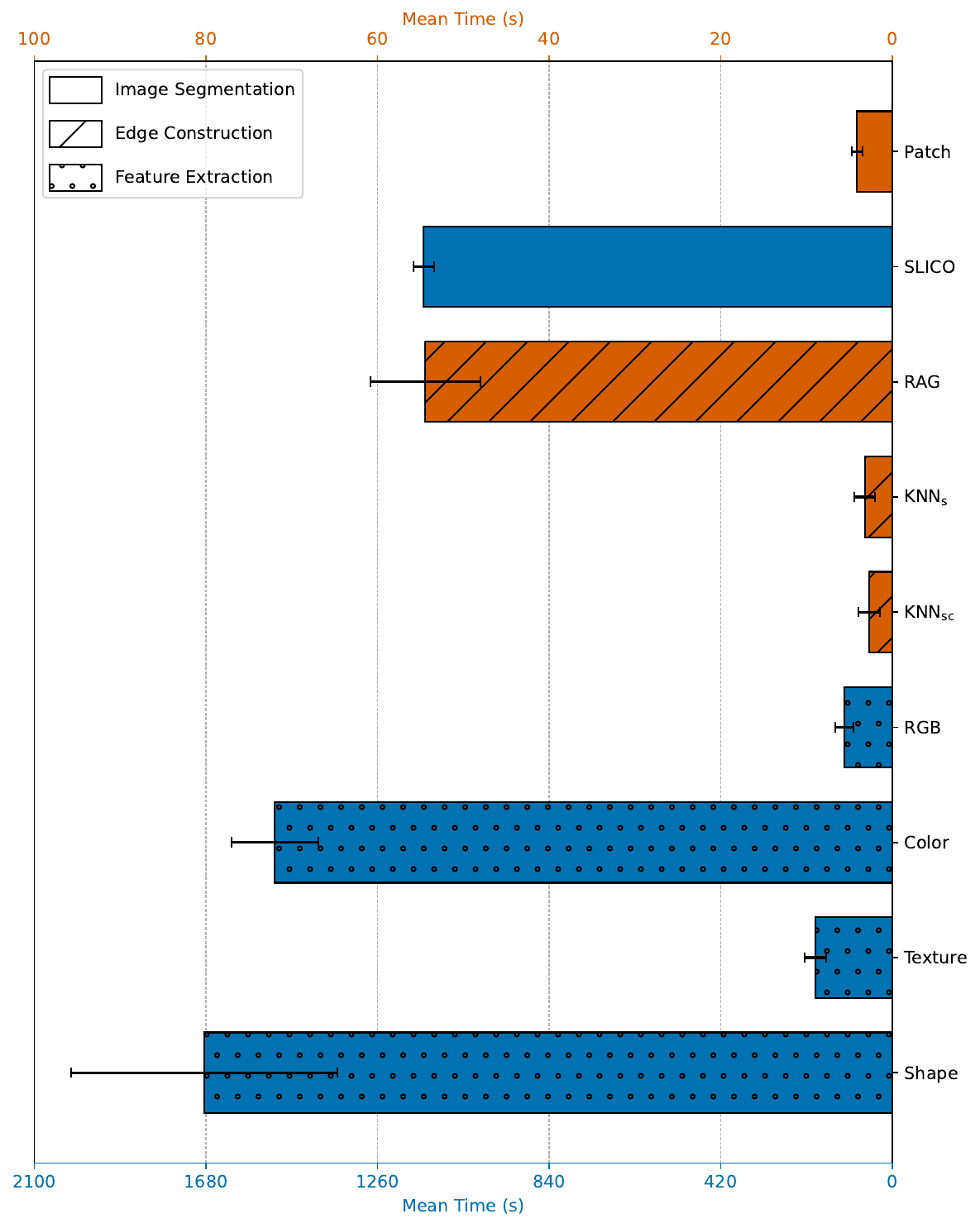}
    \caption{Image-to-graph runtimes.}
    \label{fig:preprocessing_times}
  \end{subfigure}
  \hfill
  \begin{minipage}[b]{0.48\linewidth}
    \begin{subfigure}[b]{\linewidth}
      \centering
      \includegraphics[width=\linewidth]{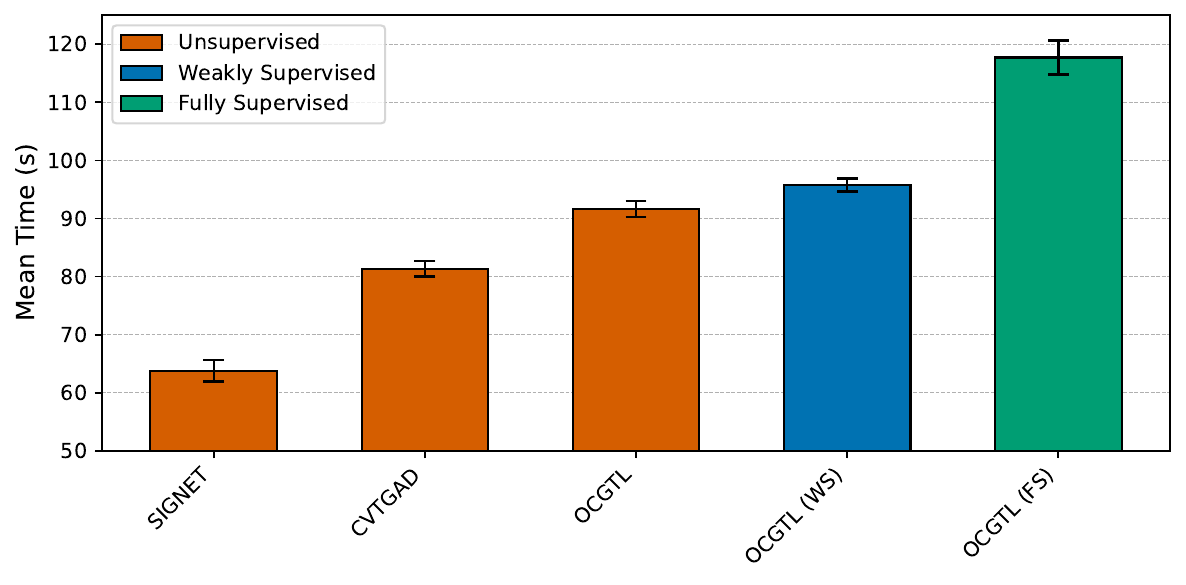}
      \caption{GLAD model runtimes.}
      \label{fig:model_runtimes}
    \end{subfigure}
    \vspace{0em} 
    
    \begin{subfigure}[b]{\linewidth}
      \centering
      \includegraphics[width=\linewidth]{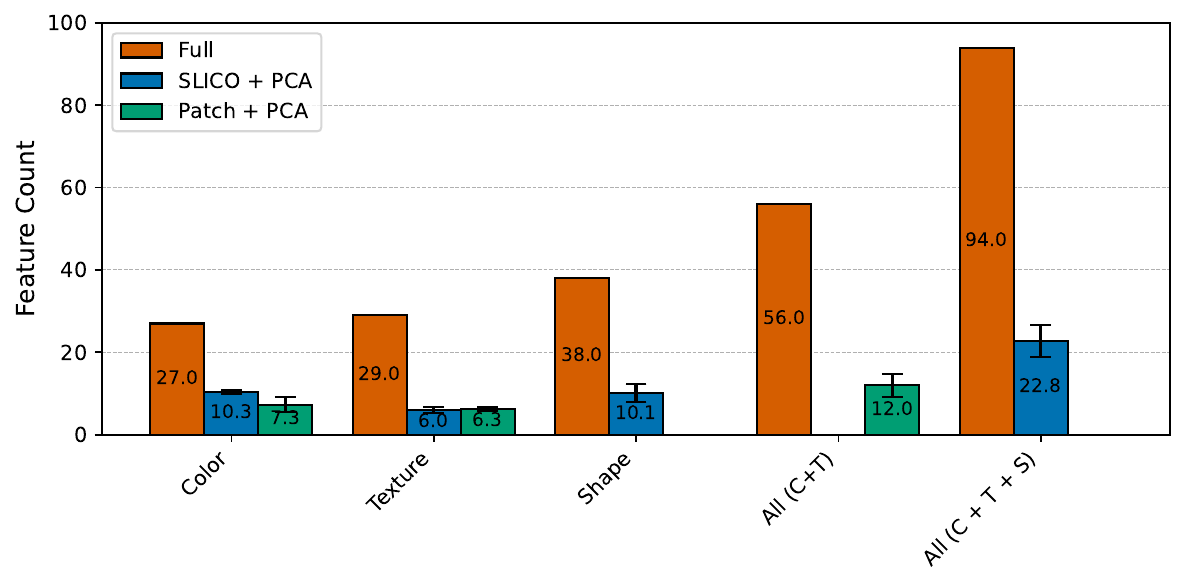}
      \caption{Feature count reduction.}
      \label{fig:feature_count}
      \vspace{0em}
    \end{subfigure}
  \end{minipage}
  \caption{(\subref{fig:preprocessing_times}) Total runtime (scaled {\textcolor{figred}{1}}:{\textcolor{figblue}{21}}) for each graph transformation on the dataset. (\subref{fig:model_runtimes}) Average model runtimes per dataset split (train \& inference) (\subref{fig:feature_count})~Feature counts for feature sets and when reduced via PCA, across all runs.}
  \label{fig:combined}
\end{figure}

In addition to anomaly detection performance, the runtime of both models and pre-processing steps is crucial for practical applicability. To contextualize these values, Appendix~\ref{app:efficiency} provides details on the execution environment.

Figure~\ref{fig:preprocessing_times} shows the average time required for each step of the image-to-graph transformation pipeline when converting the whole HAM10000 dataset to graphs. While the division into patches takes virtually no time, the segmentation via SLICO averages approx. $19$ minutes. Edge creation, performed after computing spatial and color node features, is negligible for KNN and under one minute for RAG. Node feature extraction varies more substantially: average RGB and texture features require only $2$–$3$ minutes, whereas general color and shape features take on average $25$ and $28$ minutes, respectively.

Strictly converting each image into a graph with an average of $20$ nodes and up to $94$ features (depending on the feature set) drastically reduces the feature dimensionality per image. Figure~\ref{fig:feature_count} provides an overview of the original number of features per feature set and the average reduced size after applying PCA to respective node segmentations included in the experiments. Notably, the performance reported in Section~\ref{sec:performance} was achieved with a maximum of just over $20$ features per node. This compression corresponds to a feature dimensionality per graph representation of roughly $400$ on average, compared to the original $600\times450=\num{270000}$ pixels per image.

Beyond reducing memory requirements, the image-to-graph conversion also constrains model complexity and runtime. Hence, runtimes on these graph representations are modest: the average total runtime per dataset split (train + test) across all GLAD models ranges between $1$ and approx. $2$ minutes (Figure~\ref{fig:model_runtimes}). 

\subsection{Limitations}

Due to the focus on HAM10000, our study is constrained by its focus on a single medical domain, where benign intra-class lesion variability can closely resemble genuine anomalies, and image artifacts (such as calibration markers) remain unmodeled. The competitiveness of these approaches on higher resolution images of the same domain, for example, the data set of the ISIC Challenge 2024~\cite{kurtansky2024slice}, also remains to be explored. Moreover, we have yet to directly compare our two-step graph-based pipeline with pixel-level anomaly detectors or fully end-to-end image-to-graph methods under a unified experimental setup, which would clarify their exact relative runtime efficiency and detection performance. Finally, we did not systematically explore alternative hyperparameter configurations, opting instead for sensible default values.

\section{Conclusion and Outlook}

In this work, we systematically evaluated the impact of graph-structured representations on image anomaly detection using the HAM10000 dermatoscopic image dataset. Our findings demonstrate that simple region-based graph abstractions, with drastically reduced feature dimensionality, not only significantly reduce runtime and overall dependence on pretrained models but can also achieve performance competitive to image-based models in specific domains.
While highlighting the potential of graph-based approaches, several avenues remain to investigate further application potentials and enhance performance.

Image datasets with more heterogeneous and higher-resolution content could benefit from more sophisticated node segmentation and feature extraction approaches. This approach may involve applying off‐the‐shelf, pretrained segmentation networks coupled with lightweight, pretrained deep feature extractors on each node region. Furthermore, we see promise in exploring more sophisticated edge construction methods, along with the inclusion of edge features that capture richer relationships between nodes. For instance, incorporating geometric properties, such as distances or angles between nodes or higher‐order relationships, could provide additional context that enhances anomaly detection performance.

\begin{credits}
\subsubsection{\ackname} This work has been partly supported by the Research Center Trustworthy Data Science and Security (https://rc-trust.ai), one of the Research Alliance centers within the UA Ruhr (https://uaruhr.de).

\subsubsection{\discintname}

The authors have no competing interests to declare that are relevant to the content of this article.
\end{credits}
%
%
%
\bibliographystyle{splncs04}
\bibliography{main}
%

\newpage
\appendix

\section{Image-to-Graph Transformation Details}\label{app:image_to_graph}

We applied two segmentation methods to each $450 \times 600$ HAM10000 image: dividing the image patch-based into a $4 \times 5$ grid ($20$ patches) and SLICO superpixel segmentation with $n=20$ segments. We then extracted edges via two strategies: a region adjacency graph (RAG) with connectivity $=2$ and a $k$-nearest-neighbor graph with $k=6$ (either based strictly on spatial or spatial and color similarity). 
Given two centroid coordinates $(x_i,y_i), (x_j,y_j) \in [W]\times [H]$, and their mean RGB values $(r_i, g_i, b_i), (r_j, g_j, b_j) \in \{0,\ldots,255\}^3$, the spatial distance is calculated as:
\begin{align*}
    d_{spatial}=\left\|(x_i,y_i)-(x_j,y_j)\right\|,
\end{align*}
and the spatial-color distance as:
\begin{align*}
    d_{spatial-color}&=\sqrt{\frac{(x_i-x_j)^2+(y_i-y_j)^2}{2}+\frac{(r_i-r_j)^2+(g_i-g_j)^2+(b_i-b_j)^2}{3}} \\
    &= \sqrt{\sum_{k \in \{x,y\}} \left(\frac{k_i-k_j}{\sqrt{2}}\right)^2 + \sum_{k \in \{r,g,b\}} \left(\frac{k_i-k_j}{\sqrt{3}}\right)^2} \\
    &= \left\|\left(\frac{x_i}{\sqrt{2}},\frac{y_i}{\sqrt{2}},\frac{r_i}{\sqrt{3}},\frac{g_i}{\sqrt{3}},\frac{b_i}{\sqrt{3}}\right) - \left(\frac{x_j}{\sqrt{2}},\frac{y_j}{\sqrt{2}},\frac{r_j}{\sqrt{3}},\frac{g_j}{\sqrt{3}},\frac{b_j}{\sqrt{3}}\right)\right\|
\end{align*}
The full dataset yielded roughly the same average node and edge counts, as seen in Table~\ref{tab:ham10000_metrics}.

\begin{table}[hbp!]
    \centering
    {%
    \setlength{\tabcolsep}{5pt}%
        \begin{tabular}{cccccccc}
        \toprule
        \rule{0pt}{2.6ex} & Mask & $|\mathcal{V}|_{\text{avg}}$ & $|\mathcal{V}|_{\text{min}}$ & $|\mathcal{V}|_{\text{max}}$ & $|\mathcal{E}|_{\text{avg}}^{\text{rag}}$ &$|\mathcal{E}|_{\text{avg}}^{\text{knn}_{s}}$ & $|\mathcal{E}|_{\text{avg}}^{\text{knn}_{sc}}$ \\
        \midrule 
        Patch & \xmark &  20.00 &  20 &  20 & 86.00 & 144.00 & 139.75 \\
         & \cmark &  20.00 &  20 &  20 & 86.00 & 144.00 & 140.37 \\
        \midrule
        SLICO & \xmark &  19.97 &  16 & 20 & 86.00 & 143.66 & 142.02 \\ 
         & \cmark & 19.99 & 17 & 21 & 88.24 & 142.22 & 140.49 \\
        \bottomrule
        \end{tabular}
    }
    \caption{Graph Metrics for the transformed HAM10000 dataset.}
    \label{tab:ham10000_metrics}
\end{table}

\section{GLAD Model Parameterization}\label{app:gad}

All models were trained using the same learning rate, number of epochs, batch size, and optimizer settings as detailed in Section~\ref{sec:exp_setup}. We uniformly set the hidden dimensionality to $16$ and employed two GIN message-passing layers across all architectures.
For OCGTL, we additionally evaluated two semi-supervised variants: one in which anomalies comprised $5\%$ of the overall training set and another that included all available anomalous samples from the training split.

Model-specific configurations were chosen to highlight each method’s default configuration biases. CVTGAD employs a global mean-pooling readout and maintains a $16$-dimensional embedding in its feature-view encoder, while its structure-view encoder uses a $32$-dimensional hidden space. SIGNET, by contrast, uses its default sum-pooling aggregation. Finally, OCGTL stabilizes its one-class objective via an ensemble of one reference feature extractor and five additional feature extractors.

\section{Execution Environment}\label{app:efficiency}

All experiments were performed on Ubuntu 22.04.3 LTS running on an Intel\textsuperscript{\textregistered} Xeon\textsuperscript{\textregistered} W9-3495X processor (48 cores, 96 threads; 3.4 GHz base, 4.5 GHz turbo) and a single NVIDIA RTX\textsuperscript{TM} 6000 GPU (Ada; 48 GB GDDR6 ECC; CUDA 12.0). The timings for image-to-graph transformations and GLAD models were obtained using single-core execution and single GPU utilization to ensure a fair comparison.

\section{Additional Performance Comparison}\label{app:performance}

\begin{figure}[ht!]
  \centering
  \includegraphics[width=\linewidth]{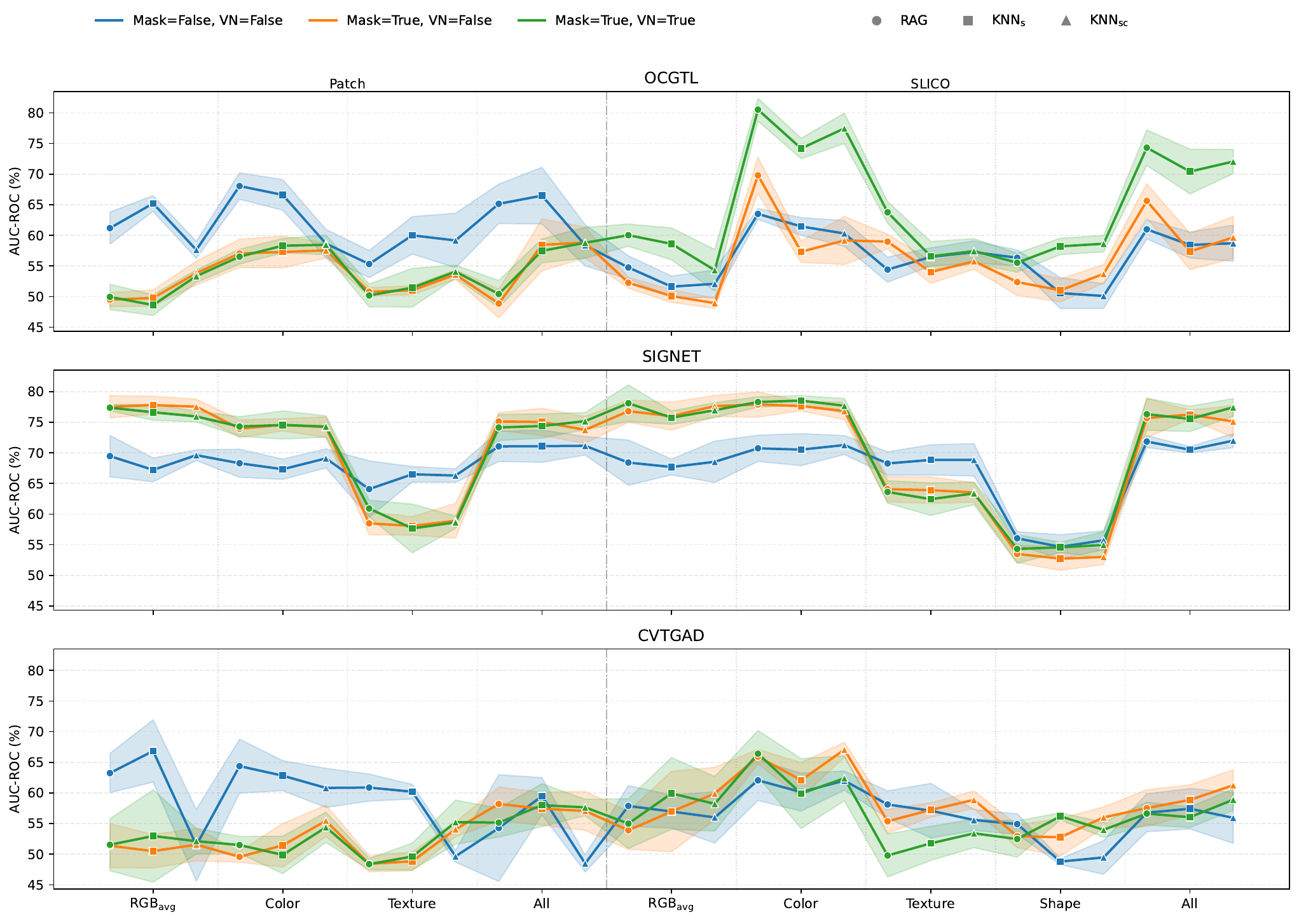}
  \caption{AUC-ROC across image-to-graph pipelines for the three unsupervised GLAD models (OCGTL, SIGNET, CVTGAD). 
  Each row shows one model with an identical x-layout (segmentation via Patch/SLICO, grouped by features).
  Lines encode the use of segmentation masks (Mask) and virtual nodes (VN); markers encode edge construction (RAG, $\mathrm{KNN}_{\mathrm{s}}$, $\mathrm{KNN}_{\mathrm{sc}}$); shaded bands indicate $\pm 1\sigma$ over splits.}
  \label{fig:rocauc_models}
\end{figure}

\begin{figure}[ht!]
  \centering
  \includegraphics[width=\linewidth]{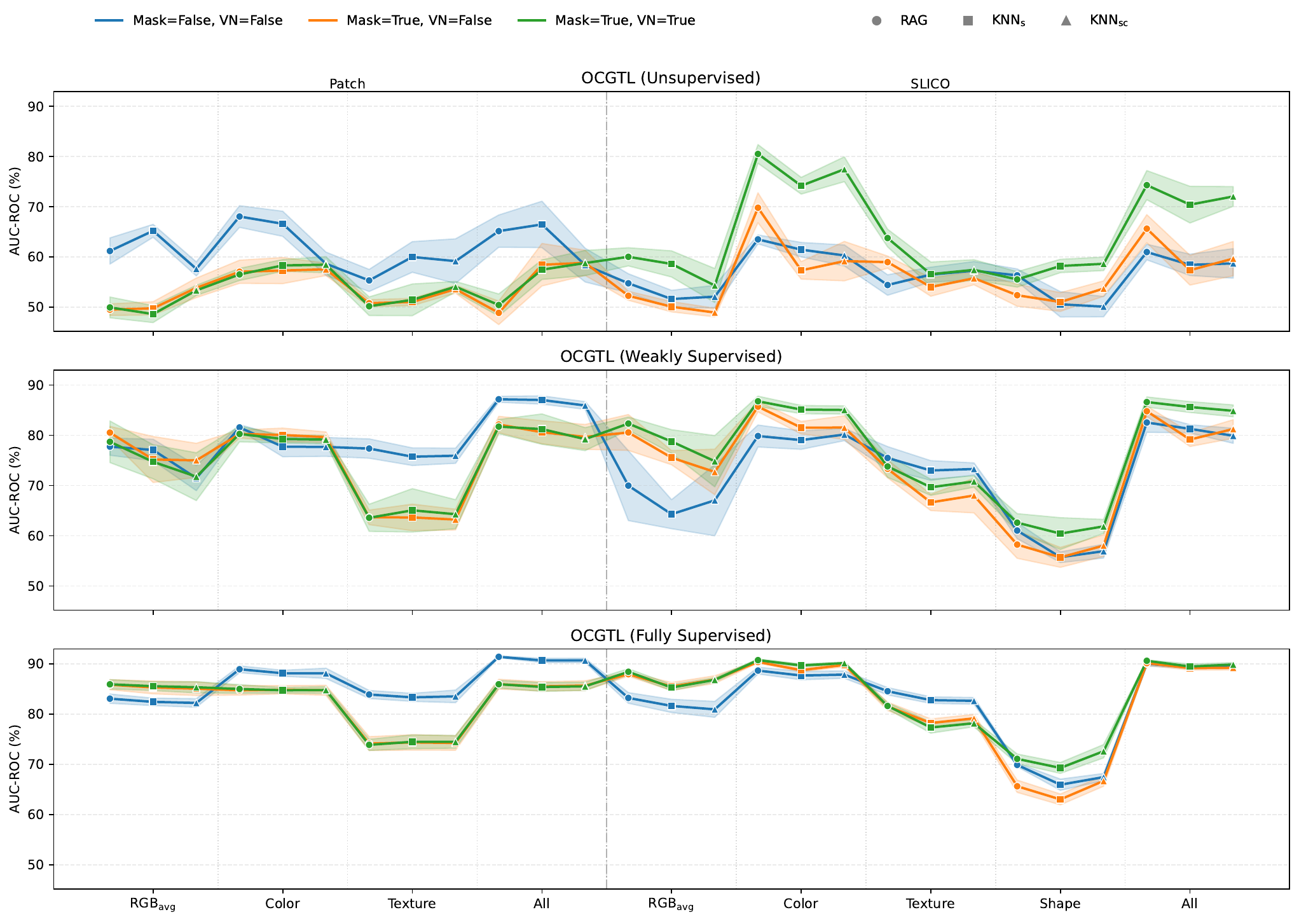}
  \caption{AUC-ROC across image-to-graph pipelines for OCGTL under three supervision regimes (Unsupervised, Weakly Supervised, Fully Supervised). 
  The visualization mirrors Fig.~\ref{fig:rocauc_models}: rows correspond to supervision levels; lines encode the use of segmentation masks (Mask) and virtual nodes (VN); markers encode edge construction; shaded bands show $\pm 1\sigma$.}
  \label{fig:rocauc_ocgtl_supervision}
\end{figure}

\begin{figure}[htbp!]
  \centering
  \includegraphics[width=\textwidth]{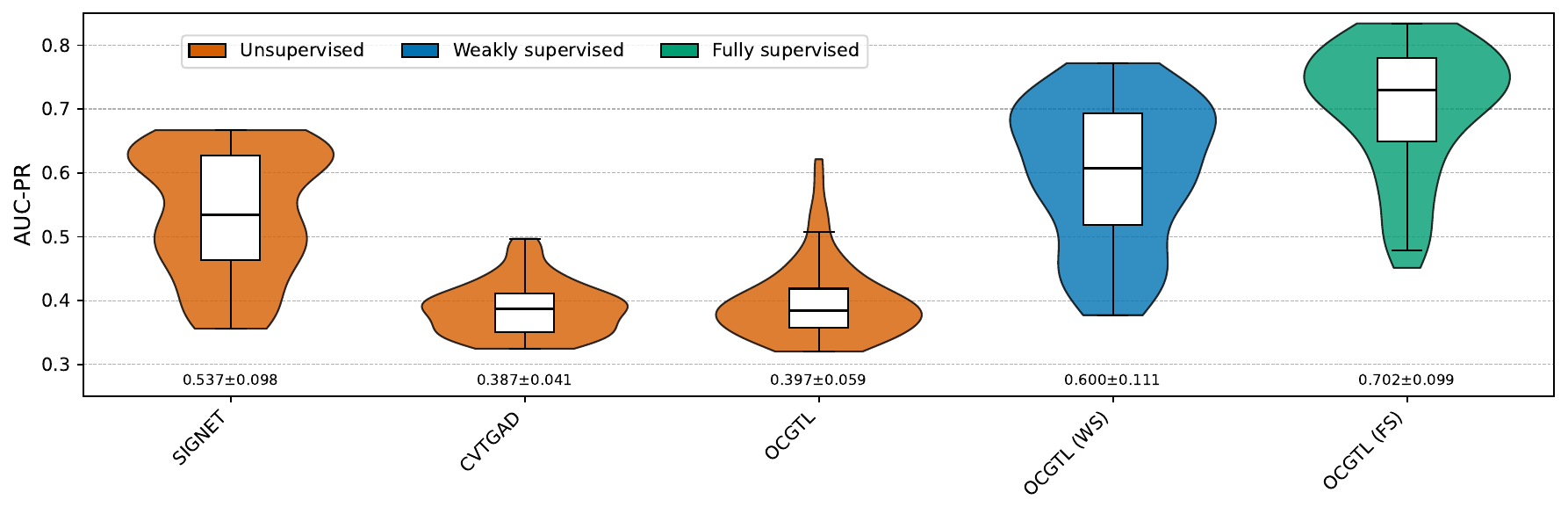}
  \caption{AUC-PR performance distribution including median, lower quartile, and upper quartile per GLAD model, as well as mean and SD, across all image-to-graph transformations.}
  \label{fig:violin_aucpr}
\end{figure}

To complement the main results, Figures~\ref{fig:rocauc_models} and \ref{fig:rocauc_ocgtl_supervision} provide an overview of the AUC-ROC performance across different image-to-graph transformation pipelines for the different GLAD models and supervision regimes, respectively. These figures visualize the relative impact of segmentation method, feature set, edge construction, and the use of segmentation masks and virtual nodes.
In addition to the AUC-ROC analysis in the main paper, Table~\ref{tab:ham10000_pr_auc} and Figure~\ref{fig:violin_aucpr} provide the corresponding AUC-PR versions of their counterparts in Section~\ref{sec:performance}. Moreover, Tables~\ref{tab:ham10000_roc_auc_all_unsupervised} and \ref{tab:ham10000_roc_auc_all_weakly_fully_supervised} present the complete AUC-ROC results, while Tables~\ref{tab:ham10000_pr_auc_all_unsupervised} and \ref{tab:ham10000_pr_auc_all_weakly_fully_supervised} list the corresponding AUC-PR values for each individual experiment.

\clearpage
\begin{table}[htbp!]
    \centering
    \begin{adjustbox}{width=\textwidth}
    \begin{tabular}{llcccccc}
        \toprule
        \multirow{2}{*}{} 
        & \multirow{2}{*}{Features} 
        & \multicolumn{3}{c}{\textsc{Patch}} 
        & \multicolumn{3}{c}{\textsc{SLICO}}  \\
        \cmidrule(lr){3-5}\cmidrule(lr){6-8}
        & 
        &  RAG     & $\text{KNN}_{\text{s}}$       & $\text{KNN}_{\text{sc}}$  
         &  RAG     & $\text{KNN}_{\text{s}}$     & $\text{KNN}_{\text{sc}}$            \\
        \midrule
        \multirow{5}{*}{\shortstack[l]{Mask \xmark \\ VN \xmark}} 
        & $\text{RGB}_{\text{avg}}$ 
            & 52.1$\pm$2.7    
            & \underline{52.3$\pm$2.2} 
            & 47.7$\pm$2.5       
            & 48.5$\pm$3.6 
            & 46.7$\pm$2.7      
            & 46.5$\pm$3.3       \\
        & Color 
            & \underline{58.1$\pm$2.2} 
            & 54.6$\pm$1.9 
            & 53.7$\pm$2.2       
            & 56.2$\pm$2.2 
            & 54.3$\pm$2.8      
            & 55.7$\pm$1.8       \\
        & Texture 
            & 50.9$\pm$2.6      
            & 51.5$\pm$1.9 
            & 50.2$\pm$2.3     
            & \underline{52.1$\pm$2.0} 
            & 51.6$\pm$3.0      
            & 51.0$\pm$1.8       \\
        & Shape 
            & ---               
            & ---               
            & ---               
            & \underline{41.7$\pm$1.3} 
            & 38.0$\pm$1.1      
            & 38.7$\pm$1.7       \\
        & All 
            & \B 59.0$\pm$2.3 
            & \underline{\B 59.3$\pm$2.4} 
            & 56.6$\pm$2.0 
            & 56.9$\pm$1.7 
            & 55.9$\pm$1.8      
            & 56.1$\pm$2.0       \\
        \midrule 
        \multirow{5}{*}{\shortstack[l]{Mask \cmark \\ VN \cmark}} 
        & $\text{RGB}_{\text{avg}}$ 
            & 53.6$\pm$2.2    
            & 53.2$\pm$2.3 
            & 52.2$\pm$2.6   
            & \underline{59.1$\pm$2.4} 
            & 56.5$\pm$3.0      
            & 54.2$\pm$3.2       \\
        & Color 
            & 55.6$\pm$1.7 
            & 55.3$\pm$2.2 
            & 55.6$\pm$1.5       
            & \underline{\B 66.9$\pm$2.5} 
            & \B 63.1$\pm$2.3      
            & \B 64.1$\pm$2.4     \\
        & Texture 
            & 42.7$\pm$2.1      
            & 42.5$\pm$2.9 
            & 43.6$\pm$2.1     
            & \underline{51.0$\pm$2.3} 
            & 46.7$\pm$2.2      
            & 47.3$\pm$1.8       \\
        & Shape 
            & ---               
            & ---               
            & ---               
            & 41.3$\pm$2.0 
            & 41.5$\pm$1.5      
            & \underline{41.9$\pm$1.7}       \\
        & All 
            & 55.7$\pm$1.9 
            & 56.9$\pm$2.7 
            & \B 57.3$\pm$1.9 
            & \underline{63.6$\pm$1.8} 
            & 61.2$\pm$2.3      
            & 62.0$\pm$1.8       \\
        \bottomrule
    \end{tabular}
    \end{adjustbox}
    \caption[Unsupervised HAM10000 PR-AUC results]{Mean AUC-PR and SD values in \% per image-to-graph transformation pipeline across all GLAD models, averaged over all splits. Best performance per column in bold, best performance per row underlined. For the features, 'All' corresponds to the complete feature set.}
    \label{tab:ham10000_pr_auc}
\end{table}

\begin{sidewaystable}[p]
    \begin{adjustbox}{width=\linewidth}
        \begin{tabular}{llcccccccccccccccccc}
        \toprule
        \multirow{3}{*}{} & 
        \multirow{3}{*}{Features} & \multicolumn{6}{c}{\large\textsc{SIGNET}} & \multicolumn{6}{c}{\large\textsc{CVTGAD}} & \multicolumn{6}{c}{\large\textsc{OCGTL}} \\
                \cmidrule(lr){3-8}\cmidrule(lr){9-14}\cmidrule(lr){15-20}
                & & \multicolumn{3}{c}{\textsc{Patch}} & \multicolumn{3}{c}{\textsc{SLICO}} & \multicolumn{3}{c}{\textsc{Patch}} & \multicolumn{3}{c}{\textsc{SLICO}} & \multicolumn{3}{c}{\textsc{Patch}} & \multicolumn{3}{c}{\textsc{SLICO}} \\
                \cmidrule(lr){3-5}\cmidrule(lr){6-8}\cmidrule(lr){9-11}\cmidrule(lr){12-14}\cmidrule(lr){15-17}\cmidrule(lr){18-20}
                & & RAG & $\text{KNN}_{\text{s}}$ & $\text{KNN}_{sc}$ & RAG & $\text{KNN}_{\text{s}}$ & $\text{KNN}_{sc}$ & RAG & $\text{KNN}_{\text{s}}$ & $\text{KNN}_{sc}$ & RAG & $\text{KNN}_{\text{s}}$ & $\text{KNN}_{sc}$ & RAG & $\text{KNN}_{\text{s}}$ & $\text{KNN}_{sc}$ & RAG & $\text{KNN}_{\text{s}}$ & $\text{KNN}_{sc}$ \\
        \midrule
        \multirow{5}{*}{\shortstack[l]{Mask \xmark \\ VN \xmark}} 
         & $\text{RGB}_{\text{avg}}$ & 69.5$\pm$3.4 & 67.2$\pm$1.9 & \underline{69.6$\pm$0.9} & 68.4$\pm$3.7 & 67.7$\pm$1.3 & 68.5$\pm$3.4 & 63.2$\pm$3.2 & \B 66.8$\pm$5.1 & 51.5$\pm$5.9 & 57.9$\pm$3.3 & 57.0$\pm$2.7 & 56.0$\pm$4.2 & 61.2$\pm$2.6 & 65.2$\pm$1.3 & 57.6$\pm$1.4 & 54.8$\pm$1.9 & 51.6$\pm$1.7 & 52.1$\pm$2.3\\
        & Color & 68.3$\pm$2.3 & 67.3$\pm$1.7 & 69.1$\pm$1.6 & 70.7$\pm$2.1 & 70.5$\pm$2.6 & \underline{71.3$\pm$1.5} & \B 64.4$\pm$4.4 & 62.8$\pm$2.4 & \B 60.8$\pm$3.2 & 62.1$\pm$3.3 & \B 60.1$\pm$3.1 & 62.0$\pm$1.6 & \B 68.1$\pm$2.2 & \B 66.6$\pm$2.5 & 58.7$\pm$2.3 & 63.5$\pm$0.9 & 61.5$\pm$1.4 & 60.3$\pm$2.1\\
        & Texture & 64.1$\pm$4.6 & 66.5$\pm$1.3 & 66.3$\pm$1.1 & 68.3$\pm$1.9 & \underline{68.9$\pm$2.5} & \underline{68.9$\pm$2.6} & 60.9$\pm$2.2 & 60.2$\pm$1.2 & 49.6$\pm$1.0 & 58.1$\pm$2.2 & 57.1$\pm$4.5 & 55.6$\pm$1.7 & 55.3$\pm$2.2 & 60.0$\pm$3.1 & \B 59.2$\pm$4.5 & 54.4$\pm$2.0 & 56.5$\pm$1.5 & 57.3$\pm$1.8\\
        & Shape & --- & --- & --- & 56.1$\pm$1.1 & 54.7$\pm$2.0 & 55.7$\pm$1.5 & --- & --- & --- & 54.9$\pm$1.7 & 48.8$\pm$0.5 & 49.5$\pm$2.7 & --- & --- & --- & \underline{56.3$\pm$1.3} & 50.6$\pm$2.5 & 50.1$\pm$2.0\\
        & All & 71.1$\pm$2.5 & 71.1$\pm$2.6 & 71.2$\pm$1.5 & 71.8$\pm$1.0 & 70.5$\pm$0.5 & \underline{72.0$\pm$1.2} & 54.3$\pm$8.7 & 59.5$\pm$3.0 & 48.5$\pm$1.4 & 56.8$\pm$3.1 & 57.4$\pm$3.3 & 55.9$\pm$4.2 & 65.1$\pm$3.2 & 66.5$\pm$4.6 & 58.4$\pm$3.4 & 61.0$\pm$1.6 & 58.4$\pm$2.1 & 58.7$\pm$2.9\\
        \midrule
        \multirow{5}{*}{\shortstack[l]{Mask \cmark \\ VN \cmark}} 
        & $\text{RGB}_{\text{avg}}$ & \B 77.4$\pm$0.5 & \B 76.6$\pm$1.3 & \B 75.9$\pm$0.9 & \underline{78.1$\pm$3.0} & 75.7$\pm$1.1 & 77.0$\pm$1.2 & 51.5$\pm$4.2 & 53.0$\pm$7.5 & 52.1$\pm$2.1 & 55.0$\pm$4.1 & 59.9$\pm$5.9 & 58.2$\pm$4.4 & 49.9$\pm$2.1 & 48.6$\pm$1.7 & 53.3$\pm$1.0 & 60.0$\pm$1.8 & 58.6$\pm$2.6 & 54.3$\pm$3.4\\
        & Color & 74.3$\pm$1.6 & 74.5$\pm$2.3 & 74.3$\pm$1.7 & \B 78.3$\pm$0.8 & \B 78.5$\pm$0.8 & \B 77.7$\pm$1.2 & 51.5$\pm$1.4 & 49.9$\pm$3.1 & 54.4$\pm$2.5 & \B 66.4$\pm$3.8 & 59.9$\pm$5.7 & \B 62.4$\pm$3.7 & 56.5$\pm$1.2 & 58.3$\pm$1.2 & 58.5$\pm$1.6 & \B \underline{80.5$\pm$1.9} & \B 74.2$\pm$1.7 & \B 77.5$\pm$2.5\\
        &Texture & 60.9$\pm$1.4 & 57.7$\pm$4.0 & 58.7$\pm$1.0 & 63.6$\pm$1.8 & 62.4$\pm$2.7 & 63.4$\pm$1.8 & 48.4$\pm$0.9 & 49.6$\pm$2.2 & 55.2$\pm$3.6 & 49.8$\pm$3.5 & 51.8$\pm$2.8 & 53.4$\pm$2.3 & 50.2$\pm$1.9 & 51.5$\pm$3.2 & 54.1$\pm$1.1 & \underline{63.8$\pm$1.8} & 56.6$\pm$2.4 & 57.4$\pm$2.0\\
        & Shape & --- & --- & --- & 54.3$\pm$2.3 & 54.6$\pm$0.8 & 55.0$\pm$2.2 & --- & --- & --- & 52.5$\pm$3.0 & 56.2$\pm$0.6 & 54.0$\pm$1.2 & --- & --- & --- & 55.5$\pm$1.5 & 58.2$\pm$1.3 & \underline{58.6$\pm$1.3}\\
        & All & 74.1$\pm$2.1 & 74.4$\pm$2.0 & 75.2$\pm$1.4 & 76.3$\pm$2.6 & 75.5$\pm$2.1 & \underline{77.4$\pm$1.4} & 55.2$\pm$2.4 & 58.0$\pm$3.5 & 57.6$\pm$1.3 & 56.6$\pm$1.9 & 56.0$\pm$1.7 & 58.9$\pm$1.8 & 50.4$\pm$2.2 & 57.5$\pm$2.0 & 58.8$\pm$2.5 & 74.3$\pm$2.9 & 70.4$\pm$3.7 & 72.0$\pm$2.0\\
        \bottomrule
        \end{tabular}
    \end{adjustbox}
    \caption[Unsupervised HAM10000 ROC-AUC results]{Unsupervised mean ROC-AUC and SD values in \% on normal class ``nv'' by GLAD method, averaged over all splits.}
    \label{tab:ham10000_roc_auc_all_unsupervised}
    \vfill
    \begin{adjustbox}{width=.48\linewidth}
        \begin{tabular}{clcccccc}
            \toprule
            \multirow{2}{*}{} & 
            \multirow{3}{*}{Features} & \multicolumn{6}{c}{\large\textsc{OCGTL (Weakly Supervised)}} \\
                    \cmidrule(lr){3-8}
                    & & \multicolumn{3}{c}{\textsc{Patch}} & \multicolumn{3}{c}{\textsc{SLICO}} \\
                    \cmidrule(lr){3-5}\cmidrule(lr){6-8}
                    & & RAG & $\text{KNN}_{\text{s}}$ & $\text{KNN}_{sc}$ & RAG & $\text{KNN}_{\text{s}}$ & $\text{KNN}_{sc}$ \\
            \midrule
            \multirow{5}{*}{\shortstack[l]{Mask \xmark \\ VN \xmark}} 
            & $\text{RGB}_{\text{avg}}$ & \underline{77.7$\pm$1.7} & 77.1$\pm$2.1 & 71.5$\pm$2.5 & 70.0$\pm$7.0 & 64.3$\pm$2.9 & 67.1$\pm$7.1\\
            & Color & \underline{81.6$\pm$0.6} & 77.7$\pm$2.0 & 77.7$\pm$1.9 & 79.9$\pm$2.2 & 79.0$\pm$1.8 & 80.2$\pm$1.2\\
            & Texture & \underline{77.4$\pm$1.9} & 75.7$\pm$1.8 & 75.9$\pm$1.5 & 75.5$\pm$2.3 & 73.0$\pm$2.0 & 73.3$\pm$1.2\\
            & Shape & --- & --- & --- & \underline{61.0$\pm$1.6} & 55.7$\pm$1.1 & 56.9$\pm$1.4\\
            & All & \B \underline{87.2$\pm$0.6} & \B 87.0$\pm$0.8 & \B 85.9$\pm$0.8 & 82.5$\pm$1.9 & 81.3$\pm$0.8 & 79.9$\pm$1.5\\
            \midrule 
            \multirow{5}{*}{\shortstack[l]{Mask \cmark \\ VN \cmark}} 
            & $\text{RGB}_{\text{avg}}$ & 78.7$\pm$4.1 & 74.7$\pm$3.4 & 71.8$\pm$4.8 & \underline{82.3$\pm$1.3} & 78.8$\pm$2.4 & 74.8$\pm$5.1\\
            & Color & 80.3$\pm$1.6 & 79.3$\pm$0.8 & 79.2$\pm$0.9 & \B \underline{86.8$\pm$1.0} & 85.1$\pm$0.8 & \B 85.0$\pm$0.8\\
            & Texture & 63.6$\pm$2.7 & 65.1$\pm$4.3 & 64.3$\pm$2.9 & \underline{73.8$\pm$2.2} & 69.7$\pm$1.6 & 70.8$\pm$1.2\\
            & Shape & --- & --- & --- & \underline{62.6$\pm$1.8} & 60.4$\pm$3.2 & 61.9$\pm$1.4\\
            & All & 81.7$\pm$1.4 & 81.2$\pm$3.0 & 79.2$\pm$2.3 & \underline{86.6$\pm$1.0} & \B 85.6$\pm$1.1 & 84.9$\pm$1.2\\
            \bottomrule
        \end{tabular}
    \end{adjustbox}
    \hfill
    \begin{adjustbox}{width=.48\linewidth}
        \begin{tabular}{clcccccc}
        \toprule
        \multirow{2}{*}{} & 
        \multirow{3}{*}{Features} & \multicolumn{6}{c}{\large\textsc{OCGTL (Fully Supervised)}} \\
                \cmidrule(lr){3-8}
                & & \multicolumn{3}{c}{\textsc{Patch}} & \multicolumn{3}{c}{\textsc{SLICO}} \\
                \cmidrule(lr){3-5}\cmidrule(lr){6-8}
                & & RAG & $\text{KNN}_{\text{s}}$ & $\text{KNN}_{sc}$ & RAG & $\text{KNN}_{\text{s}}$ & $\text{KNN}_{sc}$ \\
        \midrule
        \multirow{5}{*}{\shortstack[l]{Mask \xmark \\ VN \xmark}} 
        & $\text{RGB}_{\text{avg}}$ & 83.1$\pm$0.9 & 82.4$\pm$0.7 & 82.2$\pm$0.9 & \underline{83.2$\pm$1.1} & 81.6$\pm$1.3 & 80.9$\pm$1.6\\
        & Color & \underline{88.9$\pm$0.7} & 88.1$\pm$0.6 & 88.1$\pm$1.0 & 88.7$\pm$0.7 & 87.7$\pm$0.9 & 87.9$\pm$0.7\\
        & Texture & 83.9$\pm$0.8 & 83.3$\pm$0.8 & 83.5$\pm$1.3 & \underline{84.6$\pm$0.7} & 82.8$\pm$0.7 & 82.6$\pm$0.7\\
        & Shape & --- & --- & --- & \underline{69.9$\pm$0.4} & 65.9$\pm$1.1 & 67.4$\pm$0.7\\
        & All & \B \underline{91.4$\pm$0.2} & \B 90.7$\pm$0.4 & \B 90.7$\pm$0.4 & 90.1$\pm$0.5 & 89.3$\pm$0.2 & 89.6$\pm$0.5\\
        \midrule 
        \multirow{5}{*}{\shortstack[l]{Mask \cmark \\ VN \cmark}} 
        & $\text{RGB}_{\text{avg}}$ & 85.9$\pm$0.9 & 85.6$\pm$0.9 & 85.3$\pm$1.1 & \underline{88.4$\pm$0.6} & 85.3$\pm$0.6 & 86.8$\pm$0.4\\
        & Color & 85.0$\pm$0.9 & 84.7$\pm$0.6 & 84.8$\pm$0.7 & \B \underline{90.7$\pm$0.3} & \B 89.7$\pm$0.3 & \B 90.1$\pm$0.2\\
        & Texture & 73.9$\pm$1.1 & 74.5$\pm$1.4 & 74.5$\pm$1.3 & \underline{81.6$\pm$0.6} & 77.3$\pm$1.1 & 78.2$\pm$0.7\\
        & Shape & --- & --- & --- & 71.1$\pm$1.0 & 69.3$\pm$1.1 & \underline{72.6$\pm$1.3}\\
        & All & 85.9$\pm$0.8 & 85.4$\pm$0.9 & 85.5$\pm$0.8 & \underline{90.6$\pm$0.2} & 89.5$\pm$0.5 & 89.9$\pm$0.4\\
        \bottomrule
        \end{tabular}
    \end{adjustbox}
    \caption[Weakly and fully supervised HAM10000 ROC-AUC results]{Weakly and fully supervised (5\% and 33.1\% labeled anomalies) mean ROC-AUC and SD values in \% on normal class ``nv'' by GLAD method, averaged over all splits.}
    \label{tab:ham10000_roc_auc_all_weakly_fully_supervised}
\end{sidewaystable}

\begin{sidewaystable}[p]
    \begin{adjustbox}{width=\linewidth}
        \begin{tabular}{llcccccccccccccccccc}
        \toprule
        \multirow{3}{*}{} & 
        \multirow{3}{*}{Features} & \multicolumn{6}{c}{\large\textsc{SIGNET}} & \multicolumn{6}{c}{\large\textsc{CVTGAD}} & \multicolumn{6}{c}{\large\textsc{OCGTL}} \\
                \cmidrule(lr){3-8}\cmidrule(lr){9-14}\cmidrule(lr){15-20}
                & & \multicolumn{3}{c}{\textsc{Patch}} & \multicolumn{3}{c}{\textsc{SLICO}} & \multicolumn{3}{c}{\textsc{Patch}} & \multicolumn{3}{c}{\textsc{SLICO}} & \multicolumn{3}{c}{\textsc{Patch}} & \multicolumn{3}{c}{\textsc{SLICO}} \\
                \cmidrule(lr){3-5}\cmidrule(lr){6-8}\cmidrule(lr){9-11}\cmidrule(lr){12-14}\cmidrule(lr){15-17}\cmidrule(lr){18-20}
                & & RAG & $\text{KNN}_{\text{s}}$ & $\text{KNN}_{sc}$ & RAG & $\text{KNN}_{\text{s}}$ & $\text{KNN}_{sc}$ & RAG & $\text{KNN}_{\text{s}}$ & $\text{KNN}_{sc}$ & RAG & $\text{KNN}_{\text{s}}$ & $\text{KNN}_{sc}$ & RAG & $\text{KNN}_{\text{s}}$ & $\text{KNN}_{sc}$ & RAG & $\text{KNN}_{\text{s}}$ & $\text{KNN}_{sc}$ \\
        \midrule
        \multirow{5}{*}{\shortstack[l]{Mask \xmark \\ VN \xmark}} 
        & $\text{RGB}_{\text{avg}}$ & \underline{49.7$\pm$3.5} & 47.7$\pm$0.9 & 49.5$\pm$1.6 & 48.8$\pm$4.0 & 49.4$\pm$3.3 & 49.1$\pm$2.9 & 44.6$\pm$3.4 & \B 47.1$\pm$3.8 & 34.7$\pm$4.6 & 39.0$\pm$2.8 & 40.0$\pm$2.8 & 38.8$\pm$4.7 & 42.6$\pm$2.8 & 45.9$\pm$1.8 & 37.7$\pm$0.9 & 37.7$\pm$1.7 & 35.7$\pm$1.7 & 36.4$\pm$1.4\\
        & Color & 48.8$\pm$3.1 & 47.2$\pm$1.8 & 50.0$\pm$2.3 & 52.1$\pm$2.8 & 52.1$\pm$3.9 & \underline{53.9$\pm$2.4} & \B 48.3$\pm$3.4 & 45.4$\pm$2.6 & \B 43.3$\pm$3.5 & 43.6$\pm$3.1 & 42.3$\pm$3.6 & \B 44.8$\pm$2.4 & \B 47.4$\pm$2.7 & 44.3$\pm$1.7 & 38.7$\pm$0.8 & 43.1$\pm$1.1 & 40.5$\pm$0.9 & 39.5$\pm$1.6\\
        & Texture & 46.2$\pm$4.7 & 49.0$\pm$1.3 & 48.5$\pm$1.8 & 51.9$\pm$2.2 & \underline{53.4$\pm$3.9} & 52.0$\pm$3.8 & 40.4$\pm$1.7 & 40.4$\pm$1.2 & 34.6$\pm$0.9 & 37.9$\pm$2.1 & 40.2$\pm$4.8 & 37.9$\pm$1.5 & 36.8$\pm$2.2 & 40.7$\pm$3.0 & 40.0$\pm$3.5 & 37.2$\pm$2.3 & 37.9$\pm$1.8 & 38.2$\pm$1.0\\
        & Shape & --- & --- & --- & 38.1$\pm$1.5 & 37.1$\pm$1.2 & 37.4$\pm$2.2 & --- & --- & --- & 36.2$\pm$1.4 & 32.6$\pm$0.4 & 32.7$\pm$1.6 & --- & --- & --- & \underline{39.0$\pm$1.0} & 34.5$\pm$2.0 & 34.7$\pm$1.7\\
        & All & 53.2$\pm$2.9 & 52.9$\pm$3.7 & 53.5$\pm$2.8 & 54.7$\pm$1.2 & 53.3$\pm$0.5 & \underline{54.9$\pm$1.1} & 37.2$\pm$4.3 & 40.6$\pm$1.3 & 34.0$\pm$1.4 & 37.1$\pm$2.2 & 39.8$\pm$3.3 & 39.2$\pm$3.6 & 45.3$\pm$2.2 & \B 46.0$\pm$4.4 & 39.8$\pm$3.3 & 42.5$\pm$2.1 & 40.1$\pm$2.0 & 40.2$\pm$2.4\\
        \midrule
        \multirow{5}{*}{\shortstack[l]{Mask \cmark \\ VN \cmark}} 
        & $\text{RGB}_{\text{avg}}$ & \B \underline{65.7$\pm$1.7} & \B 65.5$\pm$2.1 & \B 64.7$\pm$2.1 & 65.4$\pm$3.1 & 63.0$\pm$2.8 & 63.0$\pm$2.5 & 33.7$\pm$2.1 & 38.0$\pm$4.1 & 34.8$\pm$2.3 & 39.6$\pm$4.3 & 42.9$\pm$5.3 & 39.4$\pm$3.4 & 32.9$\pm$1.2 & 32.0$\pm$0.6 & 34.6$\pm$1.3 & 41.5$\pm$1.7 & 41.9$\pm$3.1 & 37.4$\pm$3.1\\
        & Color & 63.2$\pm$2.8 & 62.4$\pm$3.2 & 61.5$\pm$2.3 & \B 66.2$\pm$2.7 & \B \underline{66.5$\pm$2.2} & \B 66.4$\pm$1.4 & 34.9$\pm$1.8 & 33.4$\pm$3.3 & 36.2$\pm$2.0 & \B 48.4$\pm$4.1 & \B 43.6$\pm$3.9 & 42.0$\pm$4.3 & 36.4$\pm$1.0 & 38.4$\pm$1.7 & 38.4$\pm$0.9 & \B 62.1$\pm$4.1 & \B 53.3$\pm$3.1 & \B 58.7$\pm$4.3\\
        & Texture & 43.0$\pm$1.9 & 39.7$\pm$3.6 & 40.6$\pm$2.0 & 45.2$\pm$2.1 & 44.0$\pm$3.2 & \underline{45.3$\pm$2.5} & 32.7$\pm$1.3 & 33.1$\pm$1.9 & 37.5$\pm$2.5 & 35.1$\pm$2.5 & 35.0$\pm$2.0 & 35.3$\pm$1.5 & 33.0$\pm$1.4 & 34.9$\pm$2.2 & 35.7$\pm$0.5 & 45.0$\pm$2.6 & 38.9$\pm$2.4 & 39.7$\pm$1.3\\
        & Shape & --- & --- & --- & 37.1$\pm$2.2 & 36.9$\pm$0.6 & 37.3$\pm$1.9 & --- & --- & --- & 35.1$\pm$2.4 & 37.4$\pm$0.7 & 35.3$\pm$1.8 & --- & --- & --- & 38.6$\pm$1.7 & \underline{40.9$\pm$1.7} & 40.8$\pm$1.6\\
        & All & 61.0$\pm$2.5 & 59.7$\pm$3.3 & 60.6$\pm$1.9 & 64.2$\pm$2.6 & 62.0$\pm$3.5 & \underline{64.6$\pm$2.9} & 38.6$\pm$2.9 & 41.0$\pm$3.7 & 39.4$\pm$1.7 & 40.3$\pm$1.2 & 38.0$\pm$2.0 & 39.0$\pm$1.4 & 33.3$\pm$1.7 & 41.0$\pm$2.6 & \B 42.8$\pm$3.4 & 54.6$\pm$3.9 & 50.8$\pm$4.0 & 51.8$\pm$2.4\\
        \bottomrule
        \end{tabular}
    \end{adjustbox}
    \caption[All unsupervised HAM10000 PR-AUC results]{Unsupervised mean PR-AUC and SD values in \% on normal class ``nv'' by GLAD method, averaged over all splits.}
    \label{tab:ham10000_pr_auc_all_unsupervised}
    \vfill
    \begin{adjustbox}{width=.48\linewidth}
        \begin{tabular}{clcccccc}
            \toprule
            \multirow{2}{*}{} & 
            \multirow{3}{*}{Features} & \multicolumn{6}{c}{\large\textsc{OCGTL (Weakly Supervised)}} \\
                    \cmidrule(lr){3-8}
                    & & \multicolumn{3}{c}{\textsc{Patch}} & \multicolumn{3}{c}{\textsc{SLICO}} \\
                    \cmidrule(lr){3-5}\cmidrule(lr){6-8}
                    & & RAG & $\text{KNN}_{\text{s}}$ & $\text{KNN}_{sc}$ & RAG & $\text{KNN}_{\text{s}}$ & $\text{KNN}_{sc}$ \\
            \midrule
            \multirow{5}{*}{\shortstack[l]{Mask \xmark \\ VN \xmark}} 
            & $\text{RGB}_{\text{avg}}$ & \underline{56.1$\pm$2.8} & 55.9$\pm$2.9 & 51.8$\pm$3.4 & 49.4$\pm$6.9 & 44.7$\pm$2.8 & 45.8$\pm$5.2\\
            & Color & \underline{66.9$\pm$1.0} & 59.5$\pm$2.4 & 59.5$\pm$2.7 & 64.1$\pm$3.4 & 60.7$\pm$3.9 & 63.6$\pm$1.8\\
            & Texture & \underline{62.1$\pm$2.3} & 59.6$\pm$2.0 & 59.2$\pm$1.8 & 61.3$\pm$2.0 & 57.1$\pm$3.2 & 57.6$\pm$1.3\\
            & Shape & --- & --- & --- & \underline{43.3$\pm$1.9} & 38.0$\pm$0.8 & 38.9$\pm$1.8\\
            & All & \B \underline{75.7$\pm$1.2} & \B 75.3$\pm$1.8 & \B 73.8$\pm$1.8 & 69.3$\pm$2.2 & 67.2$\pm$2.4 & 66.7$\pm$2.5\\
            \midrule 
            \multirow{5}{*}{\shortstack[l]{Mask \cmark \\ VN \cmark}} 
            & $\text{RGB}_{\text{avg}}$ & 63.5$\pm$4.1 & 58.3$\pm$3.5 & 55.3$\pm$4.6 & \underline{69.7$\pm$1.8} & 62.9$\pm$2.2 & 56.8$\pm$6.3\\
            & Color & 69.3$\pm$1.5 & 68.5$\pm$1.3 & 67.9$\pm$1.3 & \underline{75.3$\pm$1.2} & 72.3$\pm$1.9 & 72.7$\pm$1.4\\
            & Texture & 46.9$\pm$3.7 & 46.6$\pm$4.3 & 45.7$\pm$2.9 & \underline{58.8$\pm$3.1} & 52.7$\pm$1.6 & 53.2$\pm$1.6\\
            & Shape & --- & --- & --- & \underline{43.4$\pm$2.2} & 42.2$\pm$2.8 & 42.8$\pm$1.1\\
            & All & 71.0$\pm$1.2 & 69.7$\pm$2.7 & 69.2$\pm$1.5 & \B \underline{77.2$\pm$1.0} & \B 75.0$\pm$1.2 & \B 73.7$\pm$1.7\\
            \bottomrule
        \end{tabular}
    \end{adjustbox}
    \hfill
    \begin{adjustbox}{width=.48\linewidth}
        \begin{tabular}{clcccccc}
        \toprule
        \multirow{2}{*}{} & 
        \multirow{3}{*}{Features} & \multicolumn{6}{c}{\large\textsc{OCGTL (Fully Supervised)}} \\
                \cmidrule(lr){3-8}
                & & \multicolumn{3}{c}{\textsc{Patch}} & \multicolumn{3}{c}{\textsc{SLICO}} \\
                \cmidrule(lr){3-5}\cmidrule(lr){6-8}
                & & RAG & $\text{KNN}_{\text{s}}$ & $\text{KNN}_{sc}$ & RAG & $\text{KNN}_{\text{s}}$ & $\text{KNN}_{sc}$ \\
        \midrule
        \multirow{5}{*}{\shortstack[l]{Mask \xmark \\ VN \xmark}} 
        & $\text{RGB}_{\text{avg}}$ & 67.5$\pm$1.2 & 65.0$\pm$1.5 & 64.9$\pm$1.8 & \underline{67.6$\pm$2.3} & 63.5$\pm$2.6 & 62.5$\pm$2.1\\
        & Color & \underline{78.9$\pm$0.9} & 76.9$\pm$1.1 & 77.0$\pm$1.7 & 78.0$\pm$0.7 & 76.1$\pm$1.6 & 76.6$\pm$0.8\\
        & Texture & 68.9$\pm$2.0 & 67.9$\pm$2.1 & 68.6$\pm$2.1 & \underline{72.0$\pm$1.6} & 69.2$\pm$1.3 & 69.4$\pm$1.4\\
        & Shape & --- & --- & --- & \underline{51.9$\pm$0.7} & 47.8$\pm$1.0 & 49.7$\pm$1.1\\
        & All & \B \underline{83.4$\pm$1.0} & \B 81.6$\pm$1.0 & \B 81.7$\pm$0.8 & 80.7$\pm$0.7 & 79.0$\pm$0.8 & 79.6$\pm$0.4\\
        \midrule 
        \multirow{5}{*}{\shortstack[l]{Mask \cmark \\ VN \cmark}} 
        & $\text{RGB}_{\text{avg}}$ & 72.4$\pm$2.1 & 72.3$\pm$1.4 & 71.8$\pm$2.6 & \underline{79.2$\pm$1.0} & 71.7$\pm$1.4 & 74.5$\pm$0.6\\
        & Color & 73.9$\pm$1.4 & 73.5$\pm$1.3 & 74.0$\pm$1.2 & \B \underline{82.4$\pm$0.5} & 79.6$\pm$0.6 & \B 80.9$\pm$0.7\\
        & Texture & 57.9$\pm$2.1 & 58.1$\pm$2.6 & 58.4$\pm$2.5 & \underline{70.8$\pm$1.2} & 62.9$\pm$2.0 & 63.0$\pm$2.2\\
        & Shape & --- & --- & --- & 52.4$\pm$1.7 & 49.9$\pm$1.4 & \underline{53.5$\pm$2.0}\\
        & All & 74.4$\pm$1.1 & 73.2$\pm$1.4 & 74.3$\pm$0.9 & \underline{81.9$\pm$0.4} & \B 80.1$\pm$0.8 & 80.8$\pm$0.8\\
        \bottomrule
        \end{tabular}
    \end{adjustbox}
    \caption[All weakly and fully supervised HAM10000 PR-AUC results]{Weakly and fully supervised (5\% and 33.1\% labeled anomalies) mean PR-AUC and SD values in \% on normal class ``nv'' by GLAD method, averaged over all splits.}
    \label{tab:ham10000_pr_auc_all_weakly_fully_supervised}
\end{sidewaystable}

\end{document}